%% file: main.tex
\documentclass[10pt,twocolumn,letterpaper]{article}

\usepackage[pagenumbers]{cvpr} 

\input{preamble}

\usepackage{multirow}
\usepackage{indentfirst}
\usepackage{graphicx}
\usepackage{blindtext}
\usepackage{caption}
\usepackage{marvosym}
\usepackage[accsupp]{axessibility} 

\definecolor{cvprblue}{rgb}{0.21,0.49,0.74}
\usepackage[pagebackref,breaklinks,colorlinks,citecolor=cvprblue]{hyperref}

\def\paperID{12505} 
\def\confName{CVPR}
\def\confYear{2024}

\title{Co-Speech Gesture Video Generation via Motion-Decoupled Diffusion Model}

\author{Xu He$^{1}$ \quad Qiaochu Huang$^{1}$ \quad Zhensong Zhang$^{2}$ \quad Zhiwei Lin$^{1}$ \quad Zhiyong Wu$^{\textrm{\Letter},1,4}$ \\
\quad Sicheng Yang$^{1}$  \quad Minglei Li$^{3}$ \quad Zhiyi Chen$^{3}$ \quad Songcen Xu$^{2}$ \quad Xiaofei Wu$^{2}$
\\      
$^{1} $  Shenzhen International Graduate School, Tsinghua University \quad
$^{2} $ Huawei Noah’s Ark Lab \\
$^{3} $ Huawei Cloud Computing Technologies Co., Ltd \quad
$^{4} $ The Chinese University of Hong Kong\\
{\tt\small \{hex22, hqc22, lzw22, yangsc21\}@mails.tsinghua.edu.cn}
\quad {\tt\small zywu@sz.tsinghua.edu.cn} \\
{\tt\small \{zhangzhensong, liminglei29, chenzhiyi2, xusongcen, wuxiaofei2\}@huawei.com}
}

\begin{document}

\twocolumn[{%
\renewcommand\twocolumn[1][]{#1}%
\maketitle
\begin{center}
    \centering
    \captionsetup{type=figure}
    \includegraphics[width=.985\textwidth]{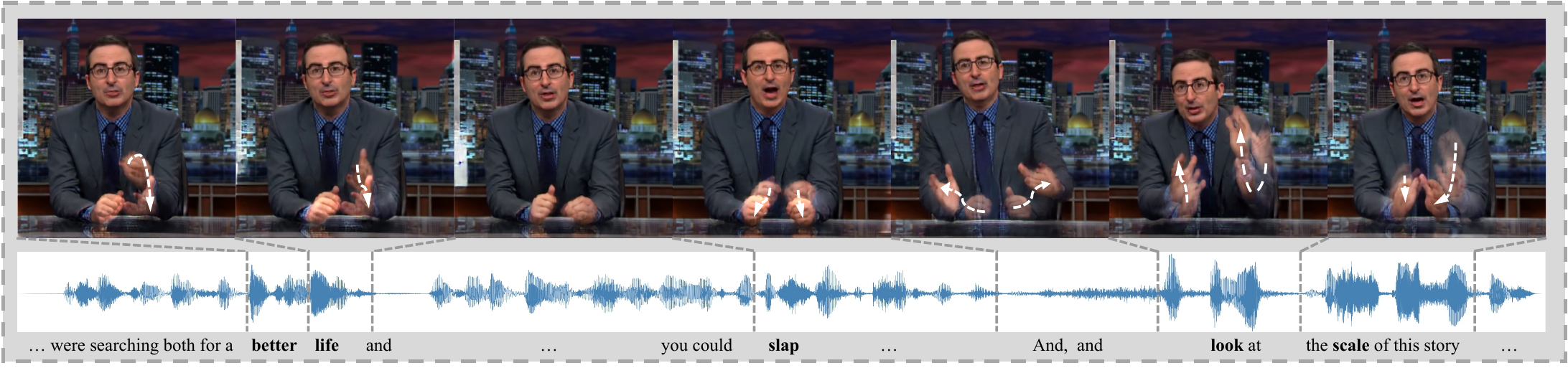}
    \captionof{figure}{\textbf{Examples of our generated gesture videos.} White dashed arrows indicate gestures corresponding to bold words.}
    \label{fig:teaser}
\end{center}%
}]

\input{sec/0_abstract}  
\vspace{-.5em}
\input{sec/1_intro}
\input{sec/2_related}

\vspace{-.5em}
\input{sec/3_method}

\input{sec/4_exp}
\vspace{-.3em}
\input{sec/5_conclusion}
\vspace{-1.3em}
\section*{Acknowledgments}
\vspace{-.1em}
This work is supported by National Natural Science Foundation of China (62076144), Shenzhen Key Laboratory of next generation interactive media innovative technology (ZDSYS20210623092001004) and Shenzhen Science and Technology Program (WDZC20220816140515001, JCYJ20220818101014030).

{
    \small
    \bibliographystyle{ieeenat_fullname}
    \bibliography{main}
}

\input{sec/X_suppl}

\end{document}

%% file: preamble.tex
\usepackage[dvipsnames]{xcolor}


%% file: sec/0_abstract.tex
\begin{abstract}
Co-speech gestures, if presented in the lively form of videos, can achieve superior visual effects in human-machine interaction. While previous works mostly generate structural human skeletons, resulting in the omission of appearance information, we focus on the direct generation of audio-driven co-speech gesture videos in this work. There are two main challenges: 1) A suitable motion feature is needed to describe complex human movements with crucial appearance information. 2) Gestures and speech exhibit inherent dependencies and should be temporally aligned even of arbitrary length. To solve these problems, we present a novel motion-decoupled framework to generate co-speech gesture videos. Specifically, we first introduce a well-designed nonlinear TPS transformation to obtain latent motion features preserving essential appearance information. Then a transformer-based diffusion model is proposed to learn the temporal correlation between gestures and speech, and performs generation in the latent motion space, followed by an optimal motion selection module to produce long-term coherent and consistent gesture videos. For better visual perception, we further design a refinement network focusing on missing details of certain areas. Extensive experimental results show that our proposed framework significantly outperforms existing approaches in both motion and video-related evaluations. Our code, demos, and more resources are available at \url{https://github.com/thuhcsi/S2G-MDDiffusion}.
\end{abstract}

%% file: sec/1_intro.tex
\section{Introduction}
\label{sec:intro}
Co-speech gestures, as a typical form of non-verbal behavior \cite{nonverbal}, convey a wealth of information and play an important role in human communication. Appropriate gestures complement human speech and thus benefit comprehension, persuasion, and credibility \cite{wilson2017hand}. Hence providing artificial agents with human-like and speech-appropriate gestures is crucial in human-machine interaction. 

To achieve this goal, several methods have been developed for automatic co-speech gesture generation, with a particular focus on deep learning techniques. 
However, they mostly aim at generating gestures as 2D/3D human skeletons. While relatively easy to generate, skeletons totally discard appearance information and create a disparity with human perception \cite{angie}. As a result, they need to be further processed for better visualization. For example, some work binds skeletons to custom virtual avatars and manually renders them using software like Blender and Maya, consuming exhaustive human labor. 
Other studies \cite{pat3,template} train independent image synthesizers \cite{pose2image} to translate skeletons into animated images, which still rely on hand-crafted annotations and yield noticeable inter-frame jitters. 

Different from previous methods that only generate skeletons, we aim to generate audio-driven co-speech gesture videos directly in a unified framework, which is challenging due to the following two reasons: First, we need to find a suitable motion feature that can describe both intricate motion trajectories and complex human appearance. A straightforward way is to design a two-stage pipeline by first generating hand-crafted and pre-defined skeletons as motion features and then synthesizing animated images with them. However, skeletons only contain positions of sparse joints and will lead to texture loss and accumulated errors, making it unsuitable for our task. Another way is to customize popular conditional video generation methods~\cite{mmdiffusion,disco,dreampose,ge2022long} to solve our problem. These methods usually encode videos into a latent space and then generate content within this space using UNet-based diffusion models~\cite{ldm1,ldm2,ldm3,ldm4}. However, they primarily concentrate on general video generation with latent features derived from VAEs lacking well-defined meaning and struggling to filter and retain necessary video information effectively. Directly applying them to videos concerning human motion results in implausible movements and missing fine-grained parts~\cite{mmdiffusion}. Second, gesture videos should be temporally aligned with the input audio even of arbitrary length, while it is still difficult to capture the inherent temporal dependencies between gestures and speech. Besides, existing video generation methods~\cite{mmdiffusion,zhou2022magicvideo} can only generate videos of fixed length, for example, 2 seconds. Generating longer consistent videos is either time-consuming or even impossible, since it requires much more computational resources.

\begin{figure*}[!ht]
  \centering
   \includegraphics[width=\linewidth]{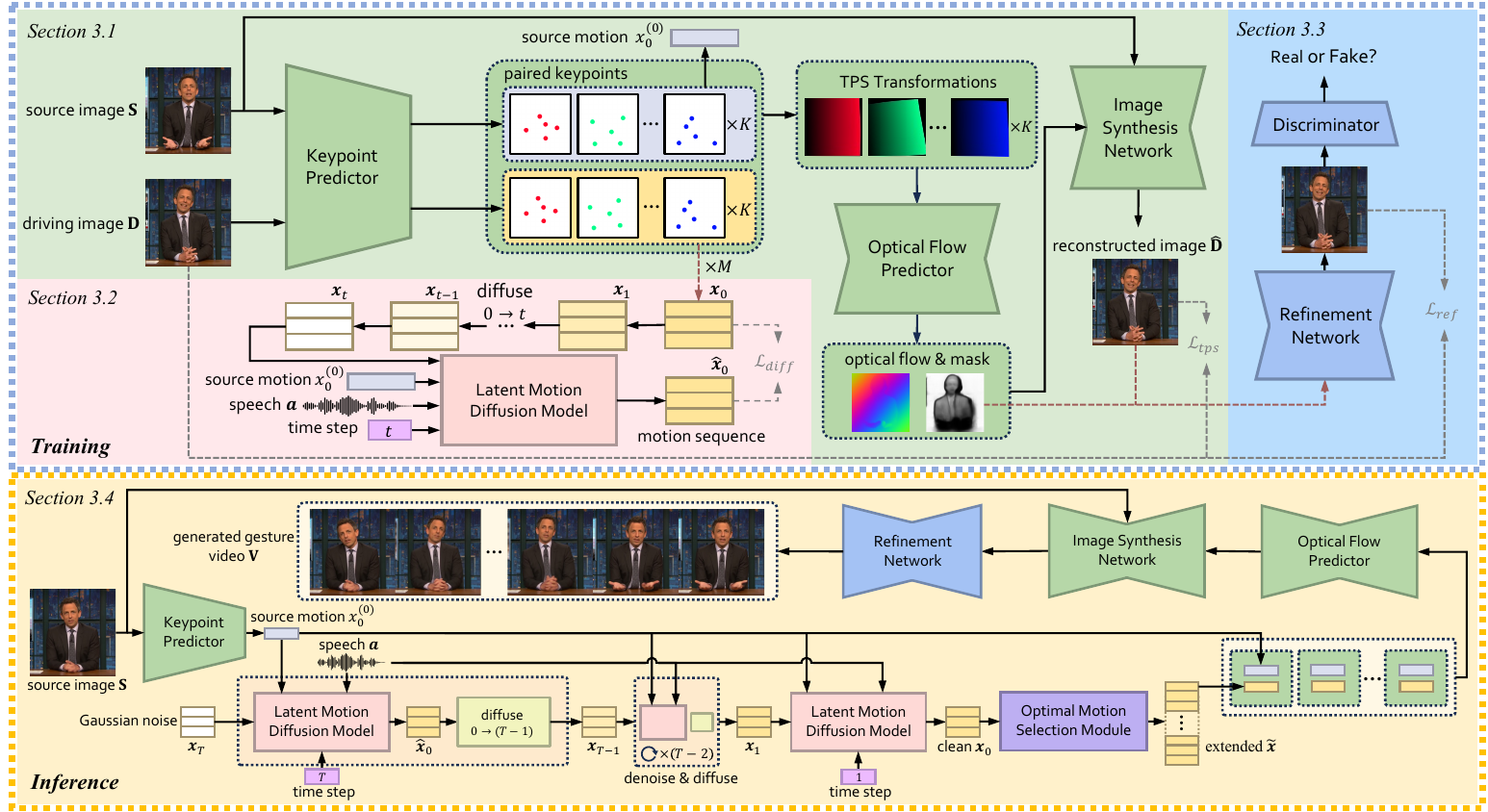}
   \caption{\textbf{Gesture video generation pipeline of our proposed framework} is composed of three core components: 1) the motion decoupling module (\textcolor[RGB]{126,152,118}{green}) extracts latent motion features from videos with TPS transformations and synthesizes image frames; 2) the latent motion diffusion model (\textcolor[RGB]{200,170,168}{pink}) generates motion features conditioned on the speech; 3) the refinement network (\textcolor[RGB]{114,136,173}{blue}) restore missing details and produce the final fine-grained video.}
   \label{fig:pipeline}
   \vspace{-.2em}
\end{figure*}

To address these challenges, in this paper, we propose a novel unified motion-decoupled framework for audio-driven co-speech gesture video generation. 
The overview of our method is shown in \cref{fig:pipeline}. 
To decouple motion from gesture videos while preserving critical appearance information of body regions, we first carefully design a thin-plate spline (TPS)~\cite{tps_transformation,tps} transformation to model first-order motion, which is nonlinear and thus flexible enough to adapt to curved human body regions. 
To be specific, we predict several groups of keypoints to generate TPS transformations, subsequently employed for estimating optical flow and guiding image warping to generate corresponding gesture video frames. Note that, gathered keypoints are considered as latent motion features, which allow for the explicit modeling of motion while maintaining a small scale, easing the burden on the generation model. 
Then we introduce a transformer-based diffusion model for generation within the latent motion space, equipped with self-attention and cross-attention modules to better capture the temporal dependency between speech and motion. To further extend the duration of generated videos, we propose an optimal motion selection module, which considers both coherence and consistency and helps to produce long-term gesture videos. 
Finally, for better visual quality, we present a UNet-like~\cite{unet} refinement network supplemented with residual blocks \cite{residual} to capture local and global information of video frames, drawing more attention to certain regions and recovering missing details of appearance and textures.

To summarize, the main contributions of our works are as follows:
\begin{itemize}
    \item We present a novel motion-decoupled framework to directly generate co-speech gesture videos in an end-to-end manner independent of hand-crafted structural human priors, where a nonlinear TPS transformation is used to extract latent motion features and ultimately guide the synthesis of gesture video frames.
    \item We design a transformer-based diffusion model on latent motion features, capturing temporal correlation between speech and gestures, which is followed by an optimal motion selection module concerning coherence and consistency. With both modules, we can generate diverse long co-speech gesture videos.
    \item We introduce a refinement network to allocate additional attention to certain areas and enhancing appearance and texture details, which is crucial for human perception.
    \item Extensive experimental results show that our framework can generate vivid, realistic, speech-matched, and long-term stable gesture videos of high quality that significantly outperform existing methods. 
\end{itemize}

%% file: sec/2_related.tex
\section{Related Works}
\label{sec:related}

\textbf{Gesture generation on human skeletons.} 
Early works consider gesture generation as an end-to-end regression task \cite{pat2,ha2g} and tend to generate averaged gestures without diversity. Subsequent insights into the many-to-many relationship between speech and gestures prompt the adoption of diverse generation methods including GANs \cite{gan}, VAEs \cite{vae}, and Flows \cite{normalizing_flow}. More currently, diffusion models excel at modeling complex data distribution and have emerged as a promising approach to generate gestures~\cite{detdiffusion,mdm,motiondiffuse,diffusestylegesture}. 
However, all these works depend on annotated datasets to generate human skeletons, including datasets labeled by pose estimators~\cite{ted,pat1,pat2} and MoCap datasets~\cite{beat,ghorbani2023zeroeggs}, suffering from error accumulation or insufficient data and totally devoid of appearance information. On the contrary, our framework generates gestures directly in the video form without relying on annotated skeleton priors.
 
\textbf{Gesture video generation.}
To date, only a few works have made initial explorations into the problem of generating gesture videos directly. Zhou \etal~\cite{graph} convert gesture video generation into a reenactment task and complete it in a rule-based way. They establish a motion graph with a reference video and search for a path matching the speech based on audio onset and a predefined keyword dictionary. However, it fails to generate novel gestures, and crafting rules is labor intensive. 
ANGIE \cite{angie} explicitly defines the problem of audio-driven co-speech gesture video generation, which utilizes an unsupervised feature, MRAA \cite{mraa}, to model body motion. Then a VQ-VAE \cite{vqvae} is leveraged to quantize common patterns, followed by a GPT-like network predicting discrete motion patterns to output gesture videos. However, as a coarse modeling of motion, MRAA is linear and fails to represent complex-shaped regions, limiting the quality of gesture videos generated by ANGIE.
Differently, we carefully design a powerful latent motion feature and a matching generation model, enabling us to generate more realistic and stable gesture videos.

\textbf{Conditional video generation.}
Another related task is conditional video generation. A variety of methods have been developed to generate videos conditioned on text \cite{lfdm}, pose \cite{dreampose,disco}, and also audio \cite{mmdiffusion}. Recently, Diffusion models are used to model video space and exhibit promising results, but their computational requirements are often substantial due to the large volume of video data. Some works \cite{ldm1,ldm2,ldm3,ldm4} adopt an auto-encoder to create a latent space for videos and subsequently, diffusion generative models can focus solely on the latent space. 
However, these methods concentrate on generating general videos. The meaning of latent features is not well-defined, which may not always preserve the desirable information such as human motion. While LaMD \cite{lamd} attempts to use two auto-encoders to separate content and motion, the separation is implicit and relies entirely on the design of the encoder network architecture. Additionally, the motion is represented as a vector without the time dimension, which may cause failure to model spatio-temporal variations in human gestures. In contrast, we design a time-aware diffusion model performing generation in a well-designed latent motion space tailored for gesture video generation and hence can generate gesture videos of high quality.

%% file: sec/3_method.tex
\section{Our Approach}
\label{sec:method}

 Given a speech audio $\boldsymbol{a}$ and a source image $\mathbf{S}$ of the speaker, our framework aims to generate an appropriate gesture video $\mathbf{V}$ (\ie an image sequence). Due to the rich connotation of gesture videos, our overall concept is to decouple and generate motion information as a bridge in the video generation process. Therefore, the pipeline can be formulated as $
\mathbf{V} \! =\! \mathcal{F}\left(\mathcal{G}\left(\mathcal{E}(\mathbf{S}),\boldsymbol{a}\right), \mathbf{S}\right)$,
where $\mathcal{E}(\cdot)$ means motion decoupling to extract the source motion feature, which will be used with the audio as conditions to facilitate the audio-to-motion conversion by the diffusion model $\mathcal{G}(\cdot)$, and finally the image synthesis and refinement network $\mathcal{F}(\cdot)$ accomplish the refined motion-to-video generation.

In the following parts, we first explain the motion decoupling module with TPS transformation, which learns latent motion features from videos and guides the source image to warp to synthesize image frames containing desired gestures (\cref{subsec:tps}). Then we elaborate the transformer-based diffusion model to perform generation within the latent motion space (\cref{subsec:diffusion}). After that, we introduce the refinement network for better visual perception which focuses more on details of specific areas (\cref{subsec:refinement}). Finally, we present the inference process of the entire framework, where the optimal motion selection module helps to produce coherent and consistent long gesture videos (\cref{subsec:inference}).

\subsection{Motion Decoupling Module with TPS}
\label{subsec:tps}
To decouple human motion from videos, a straightforward method is to extract 2D poses with off-the-shelf pose estimators \cite{openpose,dwpose}. However, as a zeroth-order model, poses completely discards appearance information around keypoints, making precise motion control and video rendering highly challenging. Furthermore, pre-training of pose estimators relies on hand-crafted 
annotations, suffering from error accumulation and often yielding jitters. The early work ANGIE \cite{angie} proposes to use MRAA \cite{mraa} consisting of mean and covariance, which is linear and fails to model regions with intricate shapes. Besides, it is inappropriate to associate covariance directly with speech. Summarizing the above, we argue that an effective representation to decouple motion is crucial for the quality of generated gesture videos and their matching with speech. Therefore, we design a motion decoupling module based on a nonlinear transformation named TPS transformation, which deals well with curving edges and hence can model the motion of various-shaped body regions. Next, we will start by introducing TPS transformation as preliminary, followed by an exposition of the entire motion decoupling module.

\textbf{TPS transformation.}
TPS transformation~\cite{tps_transformation} aims to establish the mapping $\mathcal{T}_{tps}\left(\cdot\right)$ from the origin space $\mathbf{D}$ to the deformation space $\mathbf{S}$ by utilizing known paired keypoints as control, which takes the following form:
\begin{equation}
\begin{split}
\mathcal{T}_{tps}(p)=A\left[\begin{array}{l}
p \\
1
\end{array}\right]+\sum_{i=1}^{N} w_{i} U\left(\left\|p^{\mathbf{D}}_{i}-p\right\|_{2}\right), \\
\text{s.t.} \quad \mathcal{T}_{tps}(p_{i}^{\mathbf{D}}) = p_{i}^{\mathbf{S}}, \quad i = 1,2, \ldots, N, 
\end{split}
\end{equation}
where $p=(x, y)^{\top}$ denotes coordinate. $p^\mathbf{D}_{i}$ and $p^\mathbf{S}_{i}$ are the $i^{th}$ paired keypoints from the origin and deformation space. $U(r)=r^{2} \log r^{2}$ is a radial basis function. $A \in \mathbb{R}^{2 \times 3}$ and $w_{i} \in \mathbb{R}^{2 \times 1}$ are solvable parameters as introduced in~\cite{tps_transformation}.

In our setting, given a driving and a source image corresponding to the origin space $\textbf{D}$ and the deformation space $\textbf{S}$ separately, TPS transformation can establish local connections between the two frames, which will be further used to estimate a global optical flow $\mathcal{T}(\mathbf{D})=\mathbf{S}$~\cite{tps}. It serves as the foundation for our motion decoupling module and offers two advantages: 1) as a flexible, non-linear transformation, it is suitable for modeling the motion of complex-shaped human bodies. 2) it relies solely on paired keypoints, whose movements are closely related to speech and thus can be more accurately controlled. Note that, unlike the keypoints of 2D poses only labeling certain joints, keypoints for TPS transformation come from adaptive boundary detection, involving both motion and crucial appearance information (\ie region shapes), and can be easily used for operation at pixel level and further generating video frames. 

The motion decoupling module is depicted as green in \cref{fig:pipeline}, which takes $\mathbf{S}$ and $\mathbf{D}$ as input, and outputs the constructed $\hat{\mathbf{D}}$ for end-to-end self-supervised training.

\textbf{Keypoints predictor.}
To generate TPS transformation, we first design a keypoint predictor to predict $K \times N$ keypoints, 
which will subsequently be used for producing $K$ TPS transformations with $N$ points for each. The keypoints in $\mathbf{S}$ and $\mathbf{D}$ are estimated separately and then pairwise. The collection of keypoints $\{p_{ki}\}$ is very small in scale while being capable of generating a compact optical flow to animate images. So we take it as the latent motion feature.

\textbf{Optical flow predictor.}
Now that we have $K$ TPS transformations from predicted keypoint pairs modeling local motion, we can warp the source image $\textbf{S}$ to obtain $K$ deformed images. The optical flow predictor processes the stacked deformed images and finally outputs a pixel-level optical flow indicating global motion. Following \cite{tps}, occlusion masks are also predicted, which will be fed into the image synthesis network together with the optical flow.

\textbf{Image synthesis network.}
Due to misaligned pixels and occlusions in $\mathbf{S}$ and $\mathbf{D}$, direct warping fails to generate a valid reconstructed image $\hat{\mathbf{D}}$. Hence, we propose an image synthesis network of encoder-decoder architecture, with which $\mathbf{S}$ is encoded into feature maps in different scales. The warping operation is performed on these feature maps, and occlusion masks then guide them to be masked. Subsequently, the decoder synthesizes the constructed image $\hat{\mathbf{D}}$.

\textbf{Training losses.}
From previous work \cite{fomm,mraa,tps}, we use the perceptual construction loss, equivariance loss, and warping loss to train the whole module in an unsupervised manner. The final loss is the sum of the above:
\begin{equation}
\mathcal{L}_{tps}=\mathcal{L}_{per}+\mathcal{L}_{eq}+\mathcal{L}_{warp}.
\end{equation}

For more details about training and the architecture, please refer to our supplementary material.

\subsection{Latent Motion Diffusion Model}
\label{subsec:diffusion}
Since we have decoupled motion from gesture videos, our idea is to employ a diffusion model \cite{dm1,dm2,ddpm} for generation in the latent space by denoising pure Gaussian noise. Given a real video clip, we utilize the trained keypoint predictor to obtain the keypoint sets for all frames as $\{p_{ki} \in \mathbb{R}^{2}\}^{(1:M)}$, where $M$ is the frame number. We flatten the keypoints of each frame into a $C=K \times N \times 2$-dimensioned latent motion feature and finally get a feature sequence $\boldsymbol{x}_0=x_0^{(1:M)}\in \mathbb{R}^{M \times C}$. Following \cite{ddpm}, $\boldsymbol{x}_0$ will be diffused $t$ times to get noised $\boldsymbol{x}_t$ and finally be cleaned.

\begin{figure}[t]
  \centering
   \includegraphics[width=.89\linewidth]{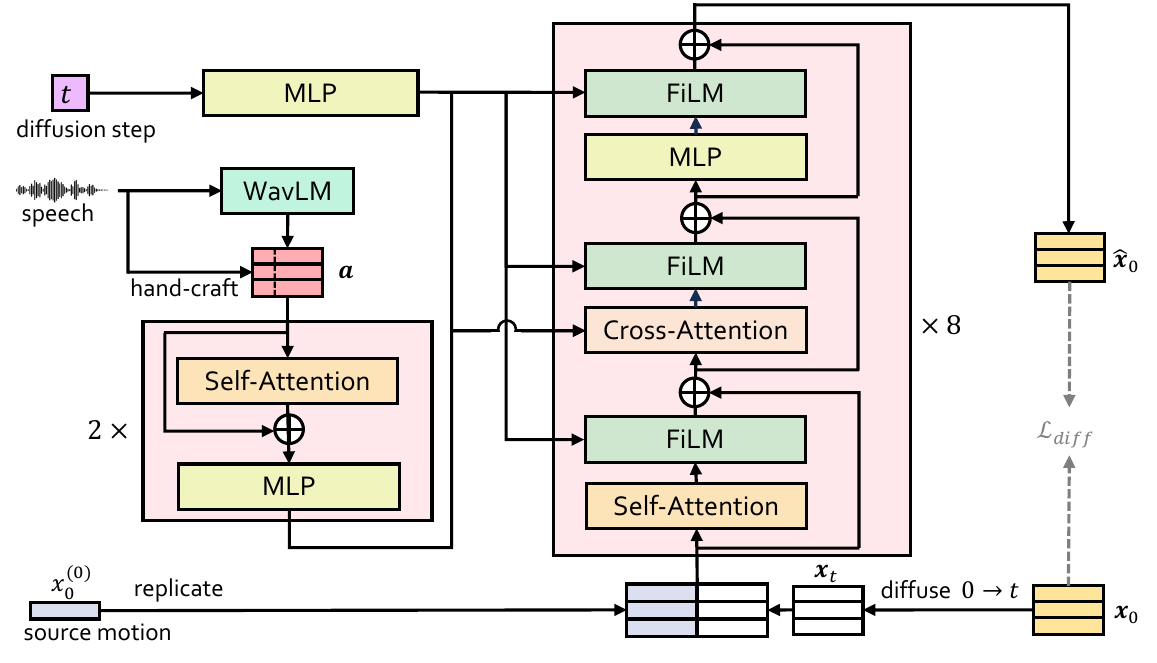}
   \caption{Noised motion features $\boldsymbol{x}_t$ are concatenated with replicated source $x_0^{(0)}$ and fed into our \textbf{transformer-based latent motion diffusion model} to predict the clean motion feature $\hat{\boldsymbol{x}}_0$ conditioned on the audio feature $\boldsymbol{a}$.
   The attention mechanism captures inherent connections between latent motion features and speech.
}
   \label{fig:diffusion}
   \vspace{-.9em}
\end{figure}

\textbf{Model.}
Per \cite{dmx0}, our diffusion model predicts the clean motion feature sequence $\hat{\boldsymbol{x}}_0$ from noised $\boldsymbol{x}_t$ given noising step $t$ and conditions $\boldsymbol{c}=\{\boldsymbol{a},x_0^{(0)}\}$, where $\boldsymbol{a}$ denotes the audio feature, and $x_0^{(0)} \in \mathbb{R}^{C} $ is the source motion feature extracted from the source image \textbf{S}, \ie the first video frame.

During training, $t$ is sampled from a uniform distribution $\mathcal{U}\{1,2,
\ldots,T\}$, and noised sequence $\boldsymbol{x}_t\in \mathbb{R}^{M \times C}$ is obtained by adding noise to $\boldsymbol{x}_0$ following DDPM \cite{ddpm}. Concerning speech audio features, \cite{diffusestylegesture} reveals that WavLM \cite{chen2022wavlm} features contain semantic information and are beneficial to the generation of co-speech motion. So we stack features generated from WavLM Large \cite{chen2022wavlm} with hand-crafted audio features to form a complete speech audio feature $\boldsymbol{a} \in \mathbb{R}^{M \times C_a}$. The former is interpolated to be aligned with the latter temporally, and $\boldsymbol{a}$ is also aligned with  $\boldsymbol{x}_t$.

The latent motion diffusion model is in a transformer-like~\cite{transformer,edge} architecture as illustrated in \cref{fig:diffusion}, which is temporally aware and well-proven for modeling motion sequences \cite{diffusestylegesture}. The encoder takes the audio feature $\boldsymbol{a}$ as input and yields hidden speech embeddings. The decoder is a transformer decoder equipped with feature-wise linear modulation (FiLM) \cite{film}. The source motion feature $x_0^{(0)}$ is replicated $M$ times to have the same temporal length as $\boldsymbol{x}_t$, which are then concatenated together and fed into the self-attention network, capturing the temporal interactions within the motion sequence. After that, speech embeddings are projected to the cross-attention layer together with the output of self-attention, which facilitates learning the inherent relationship between the motion and speech sequence.

\textbf{Training losses.}
We design the first term of loss to be common ``simple'' objective~\cite{ddpm}.
Besides, in the domain of motion generation, geometric losses \cite{geom2,mdm} are commonly used, which serve to constrain physical attributes and promote naturalness and coherence. Concerning the discussion in \cref{subsec:tps} that our latent features represent the motion, it is natural and reasonable to introduce geometric losses within the latent space. Here we use losses for velocity~\cite{mdm,siyao2022bailando} and acceleration~\cite{siyao2022bailando}.
The final training loss is as follows:
\begin{equation}
\label{eq:diffusion_loss}
\mathcal{L}_{diff}=\mathcal{L}_{simple }+\lambda_{vel} \mathcal{L}_{vel}+\lambda_{acc} \mathcal{L}_{acc}.
\end{equation}

Details can be found in the supplementary material.

\subsection{Refinement Network}
\label{subsec:refinement}
Guided by the motion features, the image synthesis network can generate speech-matched image frames according to the optical flow. However, we observe that the synthesized frames exhibit some blurs with missing details, especially in two types of regions: 1) occluded areas labeled by the occlusion masks, and 2) regions with complex textures such as hands and the face. As the image synthesis network is jointly trained with the motion decoupling module, to address this issue without disrupting the balance of motion modeling, we propose an independent refinement network.

We use a Unet-like architecture~\cite{unet} equipped with residual blocks~\cite{residual} to capture both global and local information. To draw more attention to occluded areas, the synthesized image frame is concatenated with the mask of the corresponding resolution mentioned in \cref{subsec:tps} and then fed into our refinement network. Additionally, in order to focus more on certain regions, we utilize MobileSAM \cite{mobile_sam} to segment hands and the face, and assign larger weights to both hands, face, and occluded areas in L1 reconstruction loss. Please refer to our supplementary material for more details.

\subsection{Inference}
\label{subsec:inference}
As shown in \cref{fig:pipeline}, given a source image $\mathbf{S}$ and speech as inputs, keypoints of $\mathbf{S}$ are first detected with the keypoint predictor and gathered to form the source motion feature $x_0^{(0)}$. Conditioned on $x_0^{(0)}$ and extracted audio features $\boldsymbol{a}$, we randomly sample a Gaussian noise $\boldsymbol{x}_T \in \mathbb{R}^{M\times C}$ from $\mathcal{N}\left(\mathbf{0},\mathbf{I}\right)$ and denoise it via the DDPM reverse process. At each time step $t$, the denoised sample is predicted as $\hat{\boldsymbol{x}}_{0}= \mathcal{G}(\boldsymbol{x}_t, t, \{\boldsymbol{a},x_0^{(0)}\})$ and noised back to $\boldsymbol{x}_{t-1}$. After $T$ steps, we obtain a clean sample $\boldsymbol{x}_{0}$. Repeating this procedure, we can get a consistent and coherent long sequence of motion features $\Tilde{\boldsymbol{x}}$ with a novel optimal motion selection module, detailed further below. For each frame of $\Tilde{\boldsymbol{x}}$, we can rearrange it to get $K\times N$ pairs of keypoints, producing $K$ TPS transformations along with $x_0^{(0)}$ to estimate optical flow and occlusion masks. They are then fed into the image synthesis network to generate image frames, which will go through the refinement network together with corresponding masks and finally convert into fine-grained results. All frames gather to form a complete co-speech gesture video.

\textbf{Optimal motion selection module.}
For the fact that meaningful co-speech gesture units last between 4-15 seconds~\cite{wilson2017hand,bull2006gesture}, it is crucial to generate motion feature sequences of any desired length. 
However, the transformer-based diffusion model, designed for fixed-length inputs, struggles with direct sampling of longer noise for generation due to both poor performance and high computational costs.
A naive solution is to generate fixed-length segments for concatenation, where the source motion feature $x_0^{(0)}$ is replaced by the last frame of the previous segment to ensure continuity.
However, in practice we notice that a single-frame condition cannot ensure the coherence and consistency between two segments, leading to flickers from position changes or jitters from direction changes of velocity. 

To solve this problem,  we propose an optimal motion selection module leveraging the diverse generative capability of the diffusion model, which operates solely at the inference stage. To be specific, from the second segment on, we generate $P$ candidate sequences for the same audio segment. Then a lower-better score is calculated for each candidate according to two basic assumptions: within a small time interval of a real motion sequence, 1) keypoint positions are close; 2) keypoint velocity directions are similar. Finally, the candidate motion segment with the lowest score will be selected to extend the motion sequence. Details can be found in the supplementary material. 

\begin{table*}[t]
\centering
\caption{Quantitative results on test set. Bold indicates the best and underline indicates the second. For ANGIE~\cite{angie} we reproduce the code. For MM-Diffusion~\cite{mmdiffusion} we use the officially published code. Subjective evaluation is results of MOS with 95\% confidence intervals.} 
\label{tab:comparison}
\resizebox{!}{!}{
\begin{tabular}{ccccccccc}
\hline
\multirow{2}{*}{Name}  & \multicolumn{4}{c}{Objective evaluation} & \multicolumn{4}{c}{Subjective evaluation} \\ \cline{2-9} 
& FGD $\downarrow$ &Div. $\uparrow$ & BAS $\uparrow$ & FVD $\downarrow$ & Realness $\uparrow$ & Diversity $\uparrow$ & Synchrony $\uparrow$ & Overall quality $\uparrow$\\
\hline
Ground Truth (GT) & 8.976 & 5.911 & 0.1506 & 1852.86 & 4.76$\pm$0.05 & 4.70$\pm$0.06 & 4.77$\pm$0.05 & 4.73$\pm$0.06 \\
ANGIE & 55.655 & 5.089 & \textbf{0.1504} & 2965.29 & \underline{2.07$\pm$0.08} & \underline{2.53$\pm$0.08} & \underline{2.19$\pm$0.08} & \underline{2.00$\pm$0.07} \\
MM-Diffusion & \underline{41.626} & \underline{5.189} & 0.1098 & \underline{2656.06} & 1.77$\pm$0.08 & 2.02$\pm$0.09 & 1.69$\pm$0.08 & 1.47$\pm$0.07 \\
Ours & \textbf{18.131} & \textbf{5.632} & \underline{0.1273} & \textbf{2058.19} & \textbf{3.79$\pm$0.08} & \textbf{3.91$\pm$0.07} & \textbf{3.90$\pm$0.08} & \textbf{3.77$\pm$0.07} \\
\hline
\end{tabular}}
\end{table*}

\begin{table*}[t]
\centering
\caption{Ablation study results. Bold indicates the best and underline indicates the second. `w/o' is short for `without'.}
\label{tab:ablation}
\resizebox{!}{!}{
\begin{tabular}{ccccccccc}
\hline
\multirow{2}{*}{Name}  & \multicolumn{4}{c}{Objective evaluation} & \multicolumn{4}{c}{Subjective evaluation} \\ \cline{2-9} 
& FGD $\downarrow$ &Div. $\uparrow$ & BAS $\uparrow$ & FVD $\downarrow$ & Realness $\uparrow$ & Diversity $\uparrow$ & Synchrony $\uparrow$ & Overall quality $\uparrow$\\
\hline
w/o TPS + MRAA & 288.378 & 4.625 & 0.1200 & 3034.71 & 2.59$\pm$0.09 & 2.50$\pm$0.09 & 2.59$\pm$0.09 & 1.96$\pm$0.07 \\
w/o WavLM & 37.072 & 5.344 & 0.1253 & \textbf{2053.44} & 3.44$\pm$0.08 & 3.45$\pm$0.08 & 3.43$\pm$0.08 & 3.38$\pm$0.07 \\
w/o refinement & 26.125 & 5.549 & \textbf{0.1288} & 2154.00 & \underline{3.67$\pm$0.08} & \underline{3.75$\pm$0.08} & \underline{3.74$\pm$0.07} & 3.49$\pm$0.06\\
\hline
LN Samp. & 46.055 & 4.871 & 0.1250 & 2236.72 & 2.65$\pm$0.09 & 2.25$\pm$0.09 & 2.45$\pm$0.09 & 2.70$\pm$0.09 \\
Concat. & \underline{20.964} & \underline{5.596} & 0.1250 & 2085.50 & 3.66$\pm$0.07 & 3.64$\pm$0.08 & 3.71$\pm$0.08 & \underline{3.67$\pm$0.07} \\
\hline
Ours & \textbf{18.131} & \textbf{5.632} & \underline{0.1273} & \underline{2058.19} & \textbf{3.79$\pm$0.08} & \textbf{3.91$\pm$0.07} & \textbf{3.90$\pm$0.08} & \textbf{3.77$\pm$0.07} \\
\hline
\end{tabular}}
\vspace{-.5em}
\end{table*}

%% file: sec/4_exp.tex
\section{Experiments}
\label{sec:exp}

\subsection{Experimental Settings}

\textbf{Dataset and preprocessing.}
Data of our experiments is sourced from PATS dataset \cite{pat1,pat2,pat3}, consisting of transcribed poses with aligned audios and text transcriptions, containing around 84,000 clips from 25 speakers with a mean length of 10.7s, 251 hours in total. Similar to ANGIE \cite{angie}, we perform our experiments on subsets of 4 speakers, including Jon, Kubinec, Oliver, and Seth. 
We download raw videos and audios to get clips according to PATS and conduct the following preprocessing steps:
1) Invalid clips with excessive audience applause, significant camera motion, or side views are excluded.
2) Clip lengths are limited to 4-15 seconds for meaningful gestures and resampled at 25 fps.
3) Frames are cropped with square bounding boxes, centering speakers, and resized to $256\times 256$.
4) We extend these subsets with hand-crafted onset and chromagram features and WavLM \cite{chen2022wavlm} features.
Finally, we obtain 1,200 valid clips for each speaker, randomly divided into 90\% for training and 10\% for evaluation, 4,800 in total. 

\begin{figure*}[t]
  \centering
   \includegraphics[width=.99\linewidth]{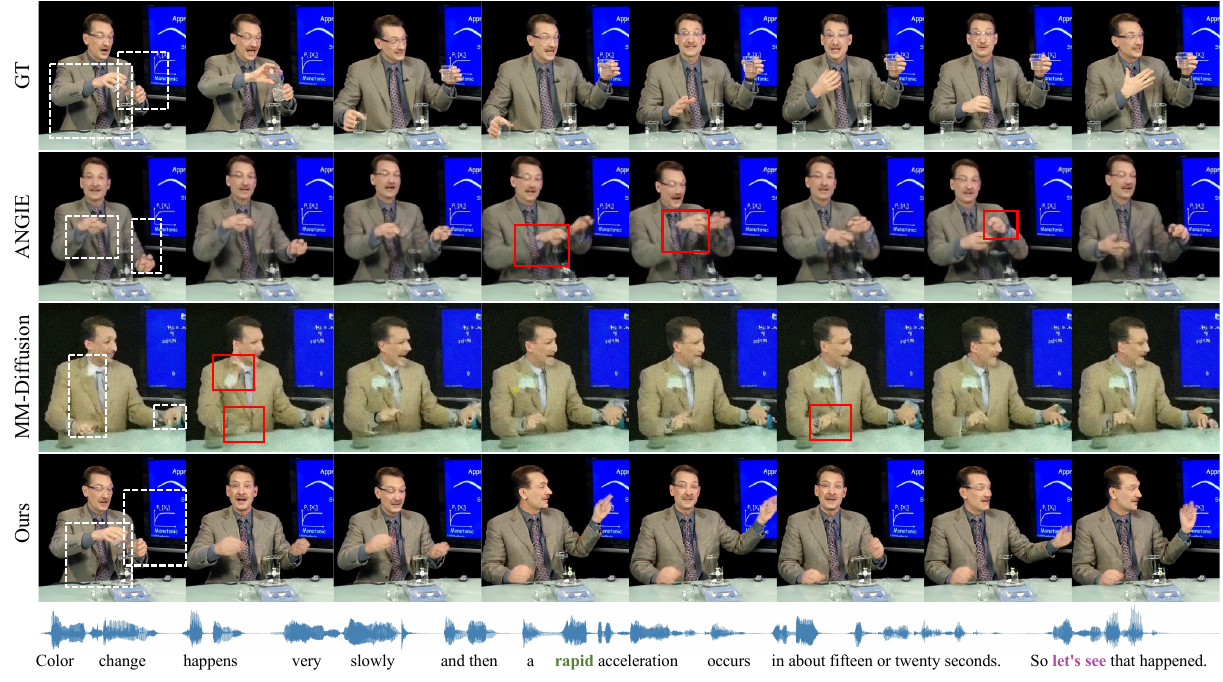}
   \caption{\textbf{Visual comparison with SOTAs.} Our method generates gestures with a broader range of motion (dashed boxes) matching both beats (green words) and semantics (purple words). Red boxes denote unrealistic gestures generated by ANGIE~\cite{angie} and MM-Diffusion~\cite{mmdiffusion}.}
   \label{fig:compare1}
   \vspace{-0.3em}
\end{figure*}

\textbf{Evaluation metrics.}
For motion-related metrics, we first extract 2D human poses with off-the-shelf pose estimator MMPose \cite{mmpose}. On this basis, we consider the quality, diversity, and alignment between gestures and speech, and choose: 1) \textbf{Fréchet Gesture Distance} (\textbf{FGD})~\cite{trimodal} to measure the distribution gap between real and generated gestures in the feature space, 2) \textbf{Diversity} (\textbf{Div.})~\cite{ha2g} which calculates feature distance between generated gestures on average. For these two metrics, we train an auto-encoder on poses from PATS. Also, we compute the average distance between closest speech beats and gesture beats as 3) \textbf{Beat Alignment Score} (\textbf{BAS}) following \cite{bas}.
For video-related metrics, we utilize 4) \textbf{Fréchet Video Distance} (\textbf{FVD})~\cite{fvd} to assess the overall quality of gesture videos. I3D \cite{wang2019i3d} classifier pre-trained on Kinetics-400 \cite{kay2017kinetics} is used to compute FVD in the feature space.

\subsection{Comparison to Existing Methods}
\label{subsec:compare}
We compare our method to: 1) the SOTA work ANGIE \cite{angie} in gesture video generation, and 2) MM-Diffusion \cite{mmdiffusion}, the SOTA work in video generation proven to be able to generate audio-driven human motion videos with experiments on AIST++ \cite{bas} human dance dataset. 

The quantitative results are reported in \cref{tab:comparison}. According to the comparison, our proposed approach significantly outperforms existing methods on motion-related metrics of FGD (56.44\%) and Diversity (8.54\%), which reveals that our motion-decoupled and diffusion-based generation framework is capable of generating realistic and diverse gestures in the motion space. Also, we achieve better performance on FVD than the best compared baseline MM-Diffusion, indicating that our method holds an advantage of ensuring the overall quality over the general audio-to-video method in gesture-specific settings. We notice that ANGIE with motion refinement tends to generate tremors synchronized with audio beat, leading to better results on BAS but at the expense of motion and visual quality. \cref{fig:compare1} presents frames of our generated videos compared with other methods, emphasizing the capacity of our method to generate videos with rich and realistic gestures matching the speech. On the contrary, limbs in ANGIE are modeled coarsely and vulnerable to abnormal deformations and absence from autoregressive error accumulation. MM-Diffusion struggles to capture body structures, leading to more or no hands.

\begin{figure}[t]
  \centering
   \includegraphics[width=.97\linewidth]{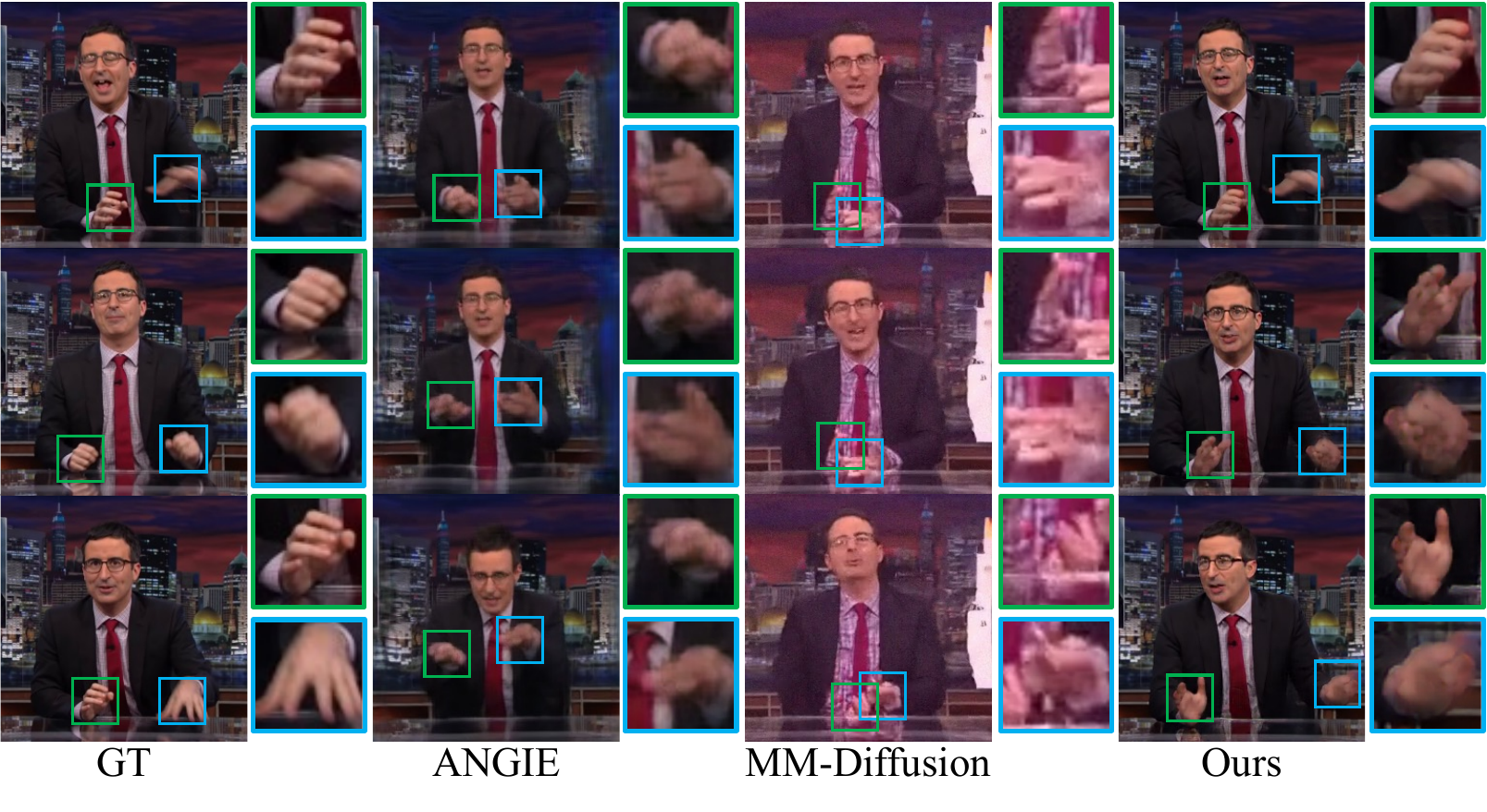}
   \caption{\textbf{
Visualization results of fine-grained hand variations.} Our generated gesture videos are more plausible and diverse.}
   \label{fig:compare2}
   \vspace{-.2em}
\end{figure}

Additionally, owing to the capability of TPS transformation to model complex-shaped regions and the close association between motion and speech established by the diffusion model, our method excels in generating precise and diverse fine-grained hand movements. As shown in \cref{fig:compare2}, directly generated videos by MM-Diffusion entirely fail to produce reasonable hand morphology. While ANGIE attempts to utilize MRAA to represent motion, this linear affine transformation coarsely models curved body regions with Gaussian distribution, resulting in hand movements presented as the translation (controlled by the mean), rotation and scaling (controlled by PCA parameters of the covariance) of an ``ellipse'' in ANGIE's results. In contrast, our method generates hand movements matching the speech, featuring intricate and plausible variations in hand shapes, which is crucial for high-quality human gestures.

\textbf{User study.}
In practice, objective metrics may not always be consistent with human subjective perceptions \cite{diffusestylegesture}, especially in the novel setting of co-speech gesture video generation. To gain further insights into the visual performance of our method, we conduct a user study to evaluate the gesture videos generated by each method alongside the ground truth. For each method, we sample 24 generated videos from the PATS test set between 3.2-12.8 seconds. 20 participants are invited to conduct the Mean Opinion Scores (MOS). Participants are asked to rate the videos in four aspects: 1) \textbf{Realness}, 2) \textbf{Diversity}, 3) \textbf{Synchrony} between speech and gestures, and 4) \textbf{Overall quality}. The first three focus on motion, while the last places more emphasis on visual perceptions. The rating scale ranges from 1 to 5 with a 1-point interval, where 1 means the poorest and 5 means the best.
The results are reported in the last four columns in \cref{tab:comparison}. Our method significantly surpasses other methods in all dimensions, which reveals that our framework can generate better gesture videos considering both motion and overall visual effects. It is noteworthy that the slight advantage of ANGIE on BAS does not translate into better gesture-speech synchrony in human subjective evaluation, where excessive tremors are not considered in sync with the speech. Please refer to the supplement for the effectiveness and robustness analysis of BAS and other objective metrics. According to the feedback from participants, ``before seeing the ground truth'', our generated gesture videos are already ``natural and well-matched to the speech enough to be mistaken as real''. Besides, there is an interesting finding that despite our emphasis on excluding irrelevant factors like textures and facial expressions in motion-related evaluations, participants express that ``when compared with the ground truth containing rich details, although generated motion is realistic, they are inevitably influenced by appearance factors''. This demonstrates that human perception of motion and appearance are interrelated. Hence generating co-speech gesture videos with visual appearance is a meaningful problem in the field of human-machine interaction.

\subsection{Ablation Study}
\label{subsec:ablation}
We conduct an ablation study to demonstrate the effectiveness of different components in our framework. The results are shown in \cref{tab:ablation}.
We explore the effectiveness of the following components: 1) the TPS-based motion decoupling module, 2) WavLM features, 3) the refinement network, and 4) the optimal motion selection module.

Supported by the results in \cref{tab:ablation}, when we replace TPS-based motion features with MRAA following ANGIE, FGD and FVD severely deteriorate by 1490\% and 47.4\%. 
When WavLM features are removed, FGD, Diversity, and BAS all deteriorate for the fact that WavLM features contain rich high-level information like semantics and emotions, crucial for driving gestures. However, WavLM brings a slight increase in FVD by 4.75, although not significantly (0.23\%), demonstrating that the positive impact of WavLM is evident in motion while having subtle effects in visual aspects.
The refinement network brings improvements in FGD, Diversity, and FVD, especially FVD decreased by 95.81 (4.4\%). Detailed analysis and visual comparisons of our ablation study can be found in the supplementary material.

For the optimal motion selection module, we replace it with two simple strategies to generate longer videos as mentioned in \cref{subsec:inference}: 1) long noise sampling (LN Samp.), and 2) direct concatenation (Concat.). According to \cref{tab:ablation}, our method equipped with the optimal motion selection module achieves the best performance across all dimensions. 

\textbf{User study.} 
Similarly, we conduct a user study for ablations as described in \cref{subsec:compare}. Results in \cref{tab:ablation} indicate that the final performance of our model decreases without any module. 
Consistent with our expectations, removing TPS has the most significant impact on the results of Realness. This reiterates the crucial significance of employing an appropriate motion feature to decouple motion. Besides, we also conduct another user study in the context of longer video generation and report the results in the supplement.

%% file: sec/5_conclusion.tex
\section{Conclusion}
\label{sec:conclusion}
In this paper, we present a novel motion-decoupled framework for co-speech gesture video generation without structural human priors. Specifically, we carefully design a nonlinear TPS transformation to obtain latent motion features, which describe motion trajectories while retaining crucial appearance information. Then, a transformer-based diffusion model is used within this latent motion space to model the intricate temporal relationship between gestures and speech, followed by an optimal motion selection module to generate diverse long gesture videos. 
Besides, a refinement network is leveraged to draw more attention to certain details and bring better visual effects. Extensive experiments demonstrate that our framework produces long-term realistic, diverse gesture videos appropriate to the given speech, and significantly outperforms existing approaches. 

%% file: sec/X_suppl.tex
\clearpage
\setcounter{page}{1}
\setcounter{section}{0}
\setcounter{figure}{0}
\setcounter{table}{0}
\setcounter{equation}{0}
\onecolumn
\centerline{\Large{\textbf{\thetitle}}\\ \vspace{0.5em}}
\centerline{\Large{Supplementary Material}}
\vspace{0.5em}
\renewcommand\thesection{\Alph{section}}
\renewcommand\thefigure{\Alph{section}\arabic{figure}}
\renewcommand\thetable{\Alph{section}\arabic{table}}
\renewcommand\theequation{\Alph{section}\arabic{equation}}

In the supplementary document, we will introduce the following contents: 1) details of \textbf{TPS transformation} (\cref{sec:supp-tps}); 2) more details of our proposed \textbf{framework} (\cref{sec:supp-framework}), including the motion decoupling module (\cref{subsec:supp-motion-decouple-module}), the latent motion diffusion model (\cref{subsec:supp-diffusion}), the refinement network (\cref{subsec:supp-refinement}), the optimal motion selection module (\cref{subsec:supp-selection}), and other implementation details (\cref{subsec:supp-implementation}); 3) the selection of \textbf{objective metrics} (\cref{sec:supp-metrics}); 4) more details and analysis of \textbf{comparison to existing methods} (\cref{sec:supp-comparison}); 5) results and analysis of the \textbf{ablation study} (\cref{sec:supp-ablation}); 6) capability of generating \textbf{long gesture videos} (\cref{sec:supp-long}); 7) \textbf{user study} details (\cref{sec:supp-user-study}); 8) analysis of the \textbf{robustness and effectiveness of objective metrics} (\cref{sec:supp-effective}); 9) \textbf{generalization ability} analysis (\cref{sec:supp-generalization}); 10) \textbf{time and resource consumption} (\cref{sec:supp-time}); 11) \textbf{limitations and future work} (\cref{sec:supp-limit}); 12) \textbf{dataset license} (\cref{sec:supp-license}).
Since more mathematical expressions are included, we choose a single-column format in this supplementary document instead of two-column for readability. All demos, code, and more resources can be found at \url{https://github.com/thuhcsi/S2G-MDDiffusion}.

\section{Details of TPS Transformation}
\label{sec:supp-tps}
In the main paper, we employ TPS transformation~\cite{tps_transformation} to establish pixel-level optical flow relying solely on sparse keypoint pairs from driving and reference images, thereby achieving precise control over the motion of human body regions. This is the foundation of our approach to decoupling motion while retaining crucial appearance information. Here we give a more detailed explanation of TPS transformation. 

TPS transformation is a type of image warping algorithm. It takes as input corresponding $N$ pairs of keypoints $(p^\mathbf{D}_i, p^\mathbf{S}_i), i=1,2,\ldots, N$ (referred to as control points) from a driving image $\mathbf{D}$ and a source image $\mathbf{S}$, and outputs a pixel coordinate mapping $\mathcal{T}_{tps}\left(\cdot\right)$ from $\mathbf{D}$ to $\mathbf{S}$ (referred to as backward optical flow). This process is grounded in the foundational assumption that the 2D warping can be emulated through a thin plate deformation model. TPS transformation seeks to minimize the energy function necessary to bend the thin plate, all while ensuring that the deformation accurately aligns with the control points, and the mathematical formulation is as follows:

\begin{equation}
\begin{split}
\min &\iint_{\mathbb{R}^{2}} \left( \left( \frac{\partial^{2} \mathcal{T}_{tps}}{\partial x^{2}} \right)^{2} + 2 \left( \frac{\partial^{2} \mathcal{T}_{tps}}{\partial x \partial y} \right)^{2} + \left( \frac{\partial^{2} \mathcal{T}_{tps}}{\partial y^{2}} \right)^{2} \right) \, dx dy, \label{eq:tps_1} \\
&\text{s.t.} \quad \mathcal{T}_{tps}(p_{i}^{\mathbf{D}}) = p_{i}^{\mathbf{S}}, \quad i = 1,2, \ldots, N,
\end{split}
\end{equation}
where $p_{i}^{\mathbf{D}}$ and $p_{i}^{\mathbf{S}}$ denotes the $i^{th}$ keypoints paired in $\mathbf{D}$ and $\mathbf{S}$. According to \cite{tps_transformation}, it can be proven that TPS interpolating function is a solution to \cref{eq:tps_1}:
\begin{equation}
\mathcal{T}_{tps}(p)=A\left[\begin{array}{l}
p \\
1
\end{array}\right]+\sum_{i=1}^{N} w_{i} U\left(\left\|p^{\mathbf{D}}_{i}-p\right\|_{2}\right), \label{eq:tps_2}
\end{equation}
where $p=(x, y)^{\top}$ is the origin coordinate in $\mathbf{D}$, and $p^\mathbf{D}_{i}$ is the $i^{th}$ keypoint in $\mathbf{D}$. $U(r)=r^{2} \log r^{2}$ is a radial basis function. Actually, $U(r)$ is the fundamental solution of the biharmonic equation \cite{biharmonic} that satisfies 
\begin{equation}
\Delta^2 U=\left(\frac{\partial^2}{\partial x^2}+\frac{\partial^2}{\partial y^2}\right)^2 U \propto \delta_{(0,0)},
\end{equation}
where the generalized function $\delta_{(0,0)}$ is defined as
\begin{align}
\delta_{(0,0)} = \begin{cases}
\infty, & \text{if } (x, y) = (0, 0) \\
0, & \text{otherwise}
\end{cases},\quad
\text{and} \iint_{\mathbb{R}^{2}} \delta_{(0,0)}(x, y) \,dx dy = 1,
\end{align}
which means that $\delta_{(0,0)}$ is zero everywhere except at the origin while having an integral equal to 1.

We use $p^{\mathbf{X}}_i=(x^{\mathbf{X}}_i, y^{\mathbf{X}}_i)^{\top}$ to denote the $i^{th}$ keypoint in image $\mathbf{X}$ (\ie~$\mathbf{D}$ or $\mathbf{S}$), and denote:
\begin{equation}
r_{ij} = \left\|p^{\mathbf{D}}_{i}-p^{\mathbf{D}}_{j}\right\|, \quad i,j=1,2,\ldots, N, \notag
\end{equation}

\begin{equation}
\begin{aligned} K= & {\left[\begin{array}{cccc}0 & U\left(r_{12}\right) & \cdots & U\left(r_{1N}\right) \\
U\left(r_{21}\right) & 0 & \cdots & U\left(r_{2 N}\right) \\ \vdots & \vdots & \ddots & \vdots \\
U\left(r_{N 1}\right) & U\left(r_{N 2}\right) & \cdots & 0\end{array}\right] }, \quad
P =\left[\begin{array}{ccc}1 & x^{\mathbf{D}}_1 & y^{\mathbf{D}}_1 \\ 1 & x^{\mathbf{D}}_2 & y^{\mathbf{D}}_2 \\ \vdots & \vdots & \vdots \\ 1 & x^{\mathbf{D}}_N & y^{\mathbf{D}}_N\end{array}\right], \notag
\end{aligned}
\end{equation}

\begin{equation}
\begin{aligned}
L & =\left[\begin{array}{cc}K & P \\ P^T & 0\end{array}\right], \quad
Y = \left[\begin{array}{ccccccc} x^{\mathbf{S}}_1 & x^{\mathbf{S}}_2 & \cdots &  x^{\mathbf{S}}_N & 0 & 0 & 0 \\
y^{\mathbf{S}}_1 & y^{\mathbf{S}}_2 & \cdots &  y^{\mathbf{S}}_N & 0 & 0 & 0 \\
\end{array}\right]^{\top}. \notag
\end{aligned}
\end{equation}

Then we can solve the affine parameters $A \in \mathcal{R}^{2 \times 3}$ and TPS parameters $w_{i} \in \mathcal{R}^{2 \times 1}$ as:

\begin{equation}
\begin{aligned}
\left[w_1, w_2, \cdots, w_N, A\right]^{\top} = L^{-1}Y.
\end{aligned}
\end{equation}

In fact, in \cref{eq:tps_2}, the first term $A\left[\begin{array}{l} 
p \\ 1 \end{array}\right]$ is an affine transformation for the alignment in the linear space of paired control points $(p^\mathbf{D}_i, p^\mathbf{S}_i)$. The second term $\sum_{i=1}^{N} w_{i} U\left(\left\|p^{\mathbf{D}}_{i}-p\right\|_{2}\right)$ introduces non-linear distortions for elevating or lowering the thin plate. With both the linear and nonlinear transformations, TPS transformation allows for precise deformation which is important to describe the motion without discarding crucial appearance information in our framework.

\section{More Details of Our Proposed Framework}
\label{sec:supp-framework}
\subsection{Motion Decoupling Module}
\label{subsec:supp-motion-decouple-module}
\textbf{Training losses.}
The motion decoupling module is trained end-to-end in an unsupervised manner. From previous works \cite{fomm,mraa,tps}, we use a pretrained VGG-19 network \cite{vgg19} to calculate the perceptual construction loss in different resolutions as the main driving loss:
\begin{equation}
\mathcal{L}_{per}=\sum_{j} \sum_{i}\left|\text{VGG19}_{i}\left(\text{DS}_{j}(\mathbf{D})\right)-\text{VGG19}_{i}(\text{DS}_{j}(\hat{\mathbf{D}}))\right|,
\end{equation}
where $\text{VGG19}_{i}$ means the $i^{th}$ layer of the VGG-19 network, while $\text{DS}_{j}$ represents $j$ downsampling operations. Also, equivariance loss is used to enhance the stability of the keypoint predictor as:
\begin{equation}
\mathcal{L}_{eq}=\left|E_{kp}(\widetilde{\mathcal{A}}({\mathbf{S}}))-\widetilde{\mathcal{A}}\left(E_{kp}(\mathbf{S})\right)\right|,
\end{equation}
where $E_{kp}$ is the keypoint predictor, and $\widetilde{\mathcal{A}}$ is a random geometric transformation operator. 

 In addition, as introduced in \cite{tps}, we also encode $\mathbf{D}$ into feature maps with the encoder of the image synthesis network, compared with warped reference feature maps to calculate the warping loss:
 
\begin{equation}
\mathcal{L}_{warp }=\sum_{i}\left|\widetilde{\mathcal{T}}^{-1}\left(E_{i}(\mathbf{S})\right)-E_{i}(\mathbf{D})\right|,
\end{equation}
where $E_{i}$ is the $i^{th}$ layer of the encoder of the image synthesis network, and $\widetilde{\mathcal{T}}^{-1}$ denotes the inverse function of the estimated optical flow, \ie~the forward optical flow from $\mathbf{R}$ to $\mathbf{D}$.

The final loss is the sum of the above terms:
\begin{equation}
\mathcal{L}_{tps}=\mathcal{L}_{per}+\mathcal{L}_{eq}+\mathcal{L}_{warp}.
\end{equation}

\subsection{Latent Motion Diffusion Model}
\label{subsec:supp-diffusion}
\textbf{Framework.}
The framework of our latent motion diffusion model is based on DDPM \cite{ddpm}, where diffusion is defined as a Markov noising process. $\boldsymbol{x}_0 \sim p(\boldsymbol{x})$ is sampled from the real data distribution (\ie~$\boldsymbol{x}_0$ is a sequence of latent motion features drawn from a real gesture video). Given constant hyper-parameters $\alpha_t \in (0,1)$ decreasing with $t$, the forward diffusion process is to add Gaussian noise to the sample:
\begin{equation}
q\left(\boldsymbol{x}_t \mid \boldsymbol{x}_{t-1}\right)=\mathcal{N}\left(\sqrt{\alpha_t} \boldsymbol{x}_{t-1},\left(1-\alpha_t\right) \mathbf{I}\right).
\label{eq:diffusion process}
\end{equation}

When the maximum time step $T$ is sufficiently large and $\alpha_t$ is small enough, we can use standard Gaussian distribution $\mathcal{N}\left(\mathbf{0},\mathbf{I}\right)$ to approximate $\boldsymbol{x}_T$. This indicates that it is possible to estimated real posterior $q\left(\boldsymbol{x}_{t-1} \mid \boldsymbol{x}_{t}\right)$ following the reverse denoising process:
\begin{equation}
    p_\theta\left(\boldsymbol{x}_{t-1} \mid \boldsymbol{x}_t\right)=\mathcal{N}\left(\boldsymbol{x}_{t-1} ; \mu_\theta\left(\boldsymbol{x}_t, t\right), \Sigma_\theta\left(\boldsymbol{x}_t, t\right)\right),
\end{equation}
where $\mu_\theta(\cdot)$ and $\Sigma_\theta(\cdot)$ mean estimating the mean and covariance via a neural network with learnable parameters $\theta$. From DDPM~\cite{ddpm}, the network predicts the noise $\epsilon_{\theta}(\boldsymbol{x}_t,t)$ and thus we can use $\mu_\theta\left(\boldsymbol{x}_t, t\right)=\frac{1}{\sqrt{\alpha_t}}\left(\boldsymbol{x}_t-\frac{1-\alpha_t}{\sqrt{1-\bar{\alpha}_t}} \epsilon_\theta\left(\boldsymbol{x}_t, t\right)\right)$ added by randomly sampled noise to estimate $\boldsymbol{x}_{t-1}$. In our context, we take speech audio and the seed motion feature of the reference frame as conditions $c$, and aim to model the conditional distribution $p_\theta(\boldsymbol{x}_0|c)$ by gradually removing the noise. Following \cite{dmx0}, we predict $\boldsymbol{x}_0$ itself instead of noise $\epsilon$. The neural network of the diffusion network can be represented as $\hat{\boldsymbol{x}}_0 = \mathcal{G}(\boldsymbol{x}_t,t,c)$.

\textbf{Training losses.}
We follow~\cite{ddpm} to use \textit{simple} objective as the first term of losses:
\begin{equation}
\mathcal{L}_{simple}=E_{\boldsymbol{x}_0 \sim q\left(\boldsymbol{x} \mid c\right), t \sim[1, T]}\left[\left\|\boldsymbol{x}_0-\mathcal{G}\left(\boldsymbol{x}_t, t, c\right)\right\|_2^2\right].
\end{equation}

Besides, as mentioned in the main paper, we use the velocity loss and the acceleration loss to constrain the physical attributes of the motion features that describe the trajectories of the keypoint movements. Velocity and acceleration are respectively defined as the first and second-order time derivatives of the keypoint positions, and here, differential methods are employed to represent derivatives~\cite{mdm,expbld,siyao2022bailando,e3d2}:
\begin{align}
\mathcal{L}_{vel} = \frac{1}{M-1} \sum_{m=1}^{M-1} \left\| \left(\boldsymbol{x}_0^{(m+1)} - \boldsymbol{x}_0^{(m)}\right) - \left(\hat{\boldsymbol{x}}_0^{(m+1)} - \hat{\boldsymbol{x}}_0^{(m)}\right) \right\|_2^2 ,
\end{align}
\begin{align}
\mathcal{L}_{acc} = \frac{1}{M-2} \sum_{m=1}^{M-2} \left\| 
\left[ \left(\boldsymbol{x}_0^{(m+2)}-\boldsymbol{x}_0^{(m+1)} \right) - \left(\boldsymbol{x}_0^{(m+1)}-\boldsymbol{x}_0^{(m)}\right) \right]  - 
\left[ \left(\hat{\boldsymbol{x}}_0^{(m+2)}-\hat{\boldsymbol{x}}_0^{(m+1)} \right) - \left(\hat{\boldsymbol{x}}_0^{(m+1)}-\hat{\boldsymbol{x}}_0^{(m)}\right) \right] \right\|_2^2 .
\end{align}

The final training loss is as follows:
\begin{equation}
\label{eq:diffusion_loss_supp}
\mathcal{L}_{diff}=\mathcal{L}_{simple }+\lambda_{vel} \mathcal{L}_{vel}+\lambda_{acc} \mathcal{L}_{acc} .
\end{equation}

\textbf{Guidance.}
Following \cite{mdm}, we train our diffusion model with classifier-free guidance. In training, we randomly mask the speech audio with a certain probability of 25\%, \ie~replacing the condition $c=\{\boldsymbol{a},x_0^{(0)}\}$ with $c_{\varnothing}=\{\varnothing,x_0^{(0)}\}$. Then, we can strike a balance between diversity and fidelity by weighting the two results with $\gamma$:
\begin{equation}
\label{eq:diffusion_guidance}
\hat{\boldsymbol{x}}_{0}=\gamma \mathcal{G}\left(\boldsymbol{x}_t, t, c\right)+(1-\gamma) \mathcal{G}\left(\boldsymbol{x}_t, t, c_{\varnothing}\right),
\end{equation}

where we can use $\gamma>1$ for extrapolating to enhance the speech condition. 

\subsection{Refinement Network}
\label{subsec:supp-refinement}
\textbf{Architecture details.}
Inspired by~\cite{inpainting}, we use a Unet-like~\cite{unet} architecture to restore missing details of synthesized image frames. In specific, we use eight ``\texttt{convolution - LeakyReLU - batch norm}'' downsampling blocks and eight ``\texttt{upsample - convolution - LeakyReLU - batch norm}'' upsampling blocks with long skip connections, which prevent the information loss during downsampling while maintaining a large receptive field. Additionally, we insert two residual blocks~\cite{residual} into the final two layers respectively, whose shallow architecture leads to a small receptive field and processes the feature maps in a sliding window manner. Simultaneously possessing large and small receptive fields enables the refinement network to capture both global and local information, thus better recovering missing details. Also, to ensure authenticity, we employ a patch-based discriminator~\cite{inpainting} trained with GAN discriminator loss $\mathcal{L}_D$ for adversarial training. Both the ground truth and refined image are converted into feature maps, with each element being discriminated as real or fake.

\textbf{Training losses.}
Firstly, we train the refinement network with the common L1 reconstruction loss. Note that, as mentioned in the main paper, we utilize MobileSAM \cite{mobile_sam} to segment hands and the face to get the masks, and assign larger weights to both hands, face, and occluded areas using the masks in L1 reconstruction loss:
\begin{equation}
\mathcal{L}_{rec}=\mathcal{L}_{valid}+
\lambda_{occ} \mathcal{L}_{occ}+
\lambda_{hand} \mathcal{L}_{hand}+
\lambda_{face} \mathcal{L}_{face}, \label{eq:refinement_weighted_L1}
\end{equation}
where we use the complement of the occlusion masks from the optical flow predictor to compute $\mathcal{L}_{valid}$.   

Then similar to~\cite{inpainting,inpaint2,inpaint4}, VGG-16~\cite{vgg19} is  used to compute the perceptual loss and style loss in the feature space as:
\begin{equation}
\mathcal{L}_{per}= \sum_{i}\left|\text{VGG16}_{i}(\mathbf{D})-\text{VGG16}_{i}(\hat{\mathbf{D}}_{ref})\right|,
\end{equation}

\begin{equation}
\mathcal{L}_{style}= \sum_{i}\left|\text{VGG16}_{i}(\mathbf{D})\cdot[\text{VGG16}_{i}(\mathbf{D})]^{\top}
-\text{VGG16}_{i}(\hat{\mathbf{D}}_{ref})\cdot [\text{VGG16}_{i}(\hat{\mathbf{D}}_{ref})]^\top \right|,
\end{equation}
where $\hat{\mathbf{D}}_{ref}$ and $\mathbf{D}$ represent the refined image frame and the real image frame respectively. $\text{VGG16}_{i}$ means the $i^{th}$ layer of the VGG-16 network, and we select $i=5,10,17$ in this work.
In addition, following \cite{inpainting,inpaint2}, the total variation (TV) loss is used as:
\begin{equation}
\mathcal{L}_{tv}=\sum_{i} \sum_{j}\left(\left| \hat{\mathbf{D}}_{ref}^{i+1, j}  -\hat{\mathbf{D}}_{ref}^{i, j} \right| +\left| \hat{\mathbf{D}}_{ref}^{i, j+1}  -\hat{\mathbf{D}}_{ref}^{i, j} \right|\right),
\end{equation}
where $\hat{\mathbf{D}}_{ref}^{i, j}$ denotes the $(i,j)$ pixel of the refined image frame. 

The final loss is the weighted sum of the above terms, along with GAN generator loss $\mathcal{L}_{G}$:
\begin{align}
\label{eq:refinement_loss}
\mathcal{L}_{ref}=\mathcal{L}_{rec}
+\lambda_{per} \mathcal{L}_{per}+
\lambda_{style}\mathcal{L}_{style}+
\lambda_{tv}\mathcal{L}_{tv}+
\lambda_{G}\mathcal{L}_{G}.
\end{align}

\subsection{Optimal Motion Selection Module}
\label{subsec:supp-selection}
We employ a segment-wise generation approach to generate motion feature sequences of arbitrary length. Inspired by~\cite{li2023finedance}, starting from the second segment, leveraging the diversity generation capability of diffusion, we generate $P$ candidates for each segment conditioned on the current audio and the end frame of the preceding segment. The scores are computed using the last five frames of the preceding segment and the first five frames of the candidate. 

Specifically, by reorganizing the motion features back into keypoint positions, we calculate two scores: 1) Position coherency score calculates the \textbf{L1 distance} between the mean positions of the preceding segment and all candidates over five frames. 2) Velocity consistency score calculates the \textbf{angle of velocity directions} in average between the preceding and candidate segments over five frames, where velocity is computed through the differential of position. These two scores are summed to obtain the final score. A lower final score indicates fewer abrupt changes in position and velocity direction between two segments, thereby reducing flickers and jitters. So the candidate segment with the lowest score is chosen to extend the motion feature sequence. The frames at the transition points are eventually filled using cubic spline interpolation.

\subsection{Other Implementation Details}
\label{subsec:supp-implementation}
We train our overall framework on four speakers jointly in three stages. 
1) For the motion decoupling module: The number of TPS transformations $K$ is set to 20, each with $N=5$ paired keypoints. 
We select ResNet18 \cite{resnet8} as the keypoint predictor for its simplicity and modify its output dimension to $20\times 5\times 2$ to match the number and dimension of keypoints. 
Following \cite{tps}, the optical flow predictor and the image synthesis network are 2D-convolution-based and produce $64\times 64$ weight maps to generate optical flow and four occlusion masks of different resolutions (32, 64, 128, 256) to synthesize image frames. We conduct training using Adam optimizer \cite{kingma2014adam} with learning rate of $2\times 10^{-4}$, $\beta_1 =0.5$, $\beta_2 =0.999$. 
2) For the latent motion diffusion model: Keypoints are gathered and unfolded into the motion feature $x \in \mathcal{R}^{200}$ for each frame. Motion features and audios are clipped to $M=80$ frames (3.2s) with stride 10 (0.4s) for training. The 35-dimension hand-crafted audio features include MFCC, constant-Q chromagram, tempogram, on-set strength and on-set beat, which are concatenated with 1024-dimension WavLM features to form $\boldsymbol{a} \in \mathcal{R}^{1059}$. For \cref{eq:diffusion_loss_supp}, we set $\lambda_{vel} = \lambda_{acc} = 1$ and use Adan optimizer \cite{xie2022adan} with learning rate of $2\times 10^{-4}$ and 0.02 weight decay for 3,000 epochs training. The maximum sampling step $T$ is 50. 3) For the refinement network: We set $\lambda_{occ}=3, \lambda_{hand}=\lambda_{face}=5$ in \cref{eq:refinement_weighted_L1}. 
Following the hyper-parameter search results in~\cite{inpaint2}, we set $\lambda_{per}=0.05, \lambda_{style}=120, \lambda_{tv}=0.1$, and $\lambda_{GAN}=0.1$ in \cref{eq:refinement_loss}. Adam optimizer \cite{kingma2014adam} with learning rate of $2 \times 10^{-4}$, $\beta_1 =0.5, \beta_2 =0.999$ is used for the refinement generator and learning rate of $4 \times 10^{-5}$ for the discriminator. The whole framework is trained on 6 NVIDIA A10 GPUs for 5 days.
In inference, $\gamma$ in \cref{eq:diffusion_guidance} is set to 2 for extrapolating to augment the speech condition. Candidate number $P$ is set to 5 for the balance between quality and inference time.

\section{Selection of Objective Metrics}
\label{sec:supp-metrics}
As a relatively unexplored task, co-speech gesture video generation lacks effective means of objective evaluation. Pioneering work ANGIE \cite{angie} simplifies the evaluation process by degrading their generation framework to 2D human skeletons before leveraging the objective metrics common in skeleton generation, which, however, only assesses the performance of the generation module in structural skeletons without considering the effectiveness of the entire framework for gesture video generation. \cite{graph} employs metrics such as LPIPS popular in image evaluation and MOVIE for video evaluation to assess gesture reenactment. However, these general visual metrics only operate in the pixel or pixel-derived feature space, neglecting the crucial body movements in gesture videos. Therefore, we propose to use both motion and video-related metrics to evaluate gesture videos. Specifically, we use \textbf{Fréchet Gesture Distance} (\textbf{FGD})~\cite{trimodal}, \textbf{Diversity} (\textbf{Div.})~\cite{div}, and \textbf{Beat Alignment Score} (\textbf{BAS})~\cite{bas} to evaluate the motion quality, and use \textbf{Fréchet Video Distance} (\textbf{FVD})~\cite{fvd} to evaluate the video quality.

\textbf{Details of motion-related metrics.} We first extract 2D human poses with off-the-shelf pose estimator MMPose \cite{mmpose}. Extracting poses after generating gesture videos avoids the degradation of our original generation framework, allowing for effective measurements of the gesture motion quality in the videos. For the feasibility of calculating metrics, we performed normalization on raw poses: 1) We preserve 13 keypoints for the upper body and 21 keypoints for each hand, 55 keypoints in total~\cite{mmpose,coco}. 2) We align the wrist points from body detection with those from hand detection. 3) For frames where the body is not detected, all keypoints are defined as centered at (128, 128). 4) For frames where hands are not detected, $21\times 2$ hand keypoints are assigned to the corresponding body wrist points.

Then, BAS can be directly computed using the audio and the normalized poses. For FGD and Diversity metrics, we follow~\cite{template} to train an auto-encoder on pose sequences from PATS train set to encode poses into a feature space. During training, pose sequences are clipped to 80 frames without overlapping. Each clip is then encoded into a 32-dimension feature. For FGD, we compute the Fréchet Distance between features of generated videos and all real videos, including both train set and test set.  For Diversity, we calculate the average Euclidean distance of generated videos in the feature space following~\cite{div}.

\section{Comparison to Existing Methods}
\label{sec:supp-comparison}
\begin{table}[t]
\centering
\caption{\textbf{Subjective evaluation results on test set with two generation schemes for MM-Diffusion.} Bold indicates the best and underline indicates the second. Results of MOS are presented with 95\% confidence intervals. Only the favorable results of MM-Diffusion-C are reported in the main paper.} 
\label{tab:supplementary_comparison}
\resizebox{!}{!}{
\begin{tabular}{ccccc}
\hline
\multirow{2}{*}{Name} & \multicolumn{4}{c}{Subjective evaluation} \\ \cline{2-5} 
& Realness $\uparrow$ & Diversity $\uparrow$ & Synchrony $\uparrow$ & Overall quality $\uparrow$\\
\hline
Ground Truth (GT) & 4.76$\pm$0.05 & 4.70$\pm$0.06 & 4.77$\pm$0.05 & 4.73$\pm$0.06 \\
ANGIE & \underline{2.07$\pm$0.08} & \underline{2.53$\pm$0.08} & \underline{2.19$\pm$0.08} & \underline{2.00$\pm$0.07} \\
MM-Diffusion-D & 1.63$\pm$0.09 & 1.98$\pm$0.09 & 1.54$\pm$0.08 & 1.46$\pm$0.08 \\
MM-Diffusion-C & 1.77$\pm$0.08 & 2.02$\pm$0.09 & 1.69$\pm$0.08 & 1.47$\pm$0.07 \\
Ours & \textbf{3.79$\pm$0.08} & \textbf{3.91$\pm$0.07} & \textbf{3.90$\pm$0.08} & \textbf{3.77$\pm$0.07} \\
\hline
\end{tabular}}
\end{table}

As stated in the main paper, we compare our method with ANGIE~\cite{angie} and MM-Diffusion~\cite{mmdiffusion}. 
For both our method and ANGIE, we use the audio and the initial frame image from PATS test set as inputs to generate corresponding 25fps gesture videos with a resolution of $256\times 256$. Given that MM-Diffusion is trained solely conditioned on audio segments to generate 1.6s video segments of 10fps, we implement it with two generation schemes: 1)  directly sampling long noise to generate videos of corresponding audio length (MM-Diffusion-D) and, 2) generating 1.6s segments for concatenation (MM-Diffusion-C). For both schemes, the generated gesture videos are resampled to 25fps. Additionally, considering that our method and MM-Diffusion-C generate fixed-length sub-clips (3.2s and 1.6s respectively) to form the full videos, both ground truth and generated videos are cropped to multiples of 3.2s for fair comparison.

User study results, including both of the two generation schemes of MM-Diffusion, are presented in \cref{tab:supplementary_comparison}. Due to space constraints, only the favorable results (MM-Diffusion-C) are reported in the main paper as ``MM-Diffusion''. It is important to note that MM-Diffusion does not use the initial frame image as a condition, thus lacking control over the appearance of the speaker in the generated videos, resulting in inconsistent speakers between concatenated segments. So, in the user study, participants are instructed to evaluate the videos generated by MM-Diffusion-C only within each 1.6s segment, neglecting the overall quality of the full-length video. This, in fact, is a lenient evaluation for disregarding the inherent limitation of MM-Diffusion in generating consistently long videos. Nonetheless, the experimental results still demonstrate the superiority of our method over MM-Diffusion in all dimensions.
Despite some setting differences, this concessive evaluation is sufficient to prove that our method surpasses MM-Diffusion when generating short segments in gesture-specific scenarios, not to mention the capability of our method to generate consistent long gesture videos. 

Constrained by computational resources and referring to the result of our user study in \cref{tab:supplementary_comparison}, only the favorable MM-Diffusion-C is used to generate 480 test videos for objective evaluation and reported as ``MM-Diffusion'' in the main paper.

\section{Ablation Study}
\label{sec:supp-ablation}
\begin{figure}[t]
  \centering
   \includegraphics[width=\linewidth]{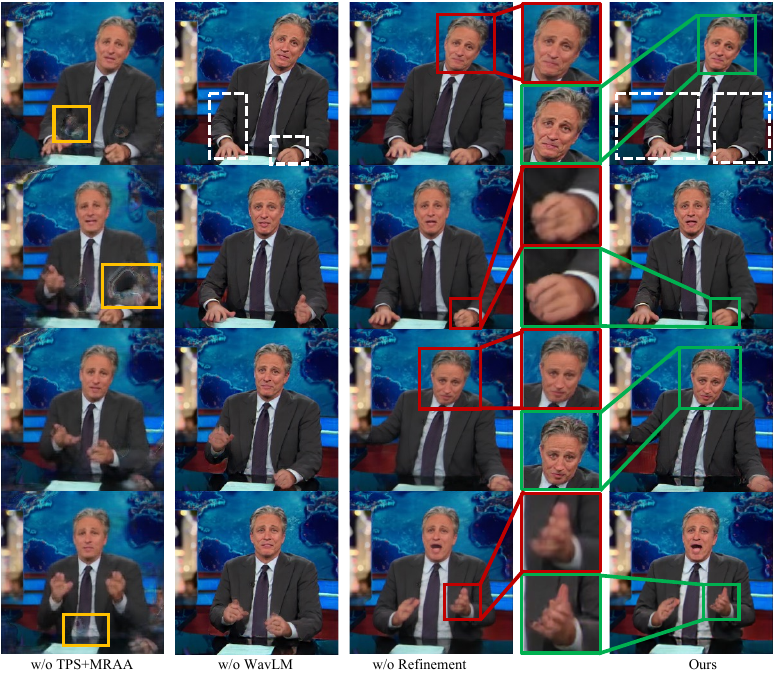}
   \caption{\textbf{Visualization results of the ablation study.} Replacing TPS with MRAA leads to ghost effects (yellow boxes). WavLM brings greater amplitude of hand motion (dashed boxes) given an impassioned speech. Refinement restores the details especially in hands and the face (red and green boxes).}
   \label{fig:ablation1}
\end{figure}
Visualization results of the ablation study are shown in \cref{fig:ablation1}, where an impassioned speech is given as the condition. From the first column, we observe that the generated videos exhibit severe ghost effects (labeled by yellow boxes) when we replace the TPS-based motion features with MRAA~\cite{mraa}. We will give an explanation in the following part. According to~\cite{mraa}, MRAA is a PCA-based affine transformation that represents motion features as the mean $\mu$ and the covariance $\Sigma$ of the probability distribution of body regions. While it is appropriate to infer $\mu$ as the region translation from speech, the interaction between speech and the region shape represented by $\Sigma$ is quite unclear. Unlike ANGIE~\cite{angie} which uses a cross-condition GPT to connect $\Sigma$ with $\mu$ and speech, our diffusion model emphasizes the interactions between speech and motion features, with less focus on relating $\Sigma$ to $\mu$. Thus the prediction of $\Sigma$ is unstable. Although we impose constraints on $\Sigma$ to be symmetric positive definite using Cholesky decomposition as mentioned in~\cite{angie} for valid gestures, it still tends to output near-singular matrices, resulting severe errors in heatmaps for the estimation of the optical flow and occlusion masks. This, in turn, causes undesirable visual effects.

The second column shows the results of removing WavLM~\cite{chen2022wavlm} features with only hand-crafted audio features used. Given an impassioned speech, the generated gestures with WavLM display greater amplitude and heightened intensity, because WavLM contains rich high-level information such as emotions and semantics~\cite{diffusestylegesture}. The final three columns of \cref{fig:ablation1} show that textures are restored after refinement, especially in hands and the face.

Please refer to
our \href{https://github.com/thuhcsi/S2G-MDDiffusion}{homepage} for more visualization results of comparison with other methods and the ablation study.

\section{Capability of Generating Long Gesture Videos}
\label{sec:supp-long}

\begin{table}[t]
\centering
\caption{\textbf{Results of generating long gesture videos.} Bold indicates the best and underline indicates the second.} 
\label{tab:supplementary_long}
\resizebox{!}{!}{
\begin{tabular}{cc}
\hline
Name & Effective duration $\uparrow$ \\ \cline{1-2} 
Ground Truth (GT) & 27.8s\\
ANGIE & 4.1s\\
LN Samp. & 3.5s\\
Concat. & \underline{15.9s}\\
Ours & \textbf{21.0s}\\
\hline
\end{tabular}}
\end{table}

To better assess the effectiveness of the optimal motion selection module and the capability of our framework to generate long gesture videos, we conduct another user study following~\cite{li2023finedance}. We sample 10 long audios from the original PATS dataset as conditions to generate videos of 28s, and compare the generated results of 1) our complete framework, 2) long noise sampling (LN Samp.), 3) direct concatenation (Concat.), 4) ANGIE, and 5) the ground truth. 20 participants are asked to evaluate the effective duration of the videos, \ie~to decide how many seconds of the videos are effective. The average effective duration for each method is shown in \cref{tab:supplementary_long}. The results show that, although based on an easy-to-make hand-crafted rule, the optimal motion selection module benefits our method to generate longer videos with better coherency and consistency compared to only seed motion used and other methods. Directly sampling long noise and the autoregressive generation approach of ANGIE both face challenges in generating effective videos over 10 seconds.

\section{Details of User Study}
\label{sec:supp-user-study}
\begin{figure*}
  \centering
  \begin{subfigure}{.9\linewidth}
    \includegraphics[width=\linewidth]{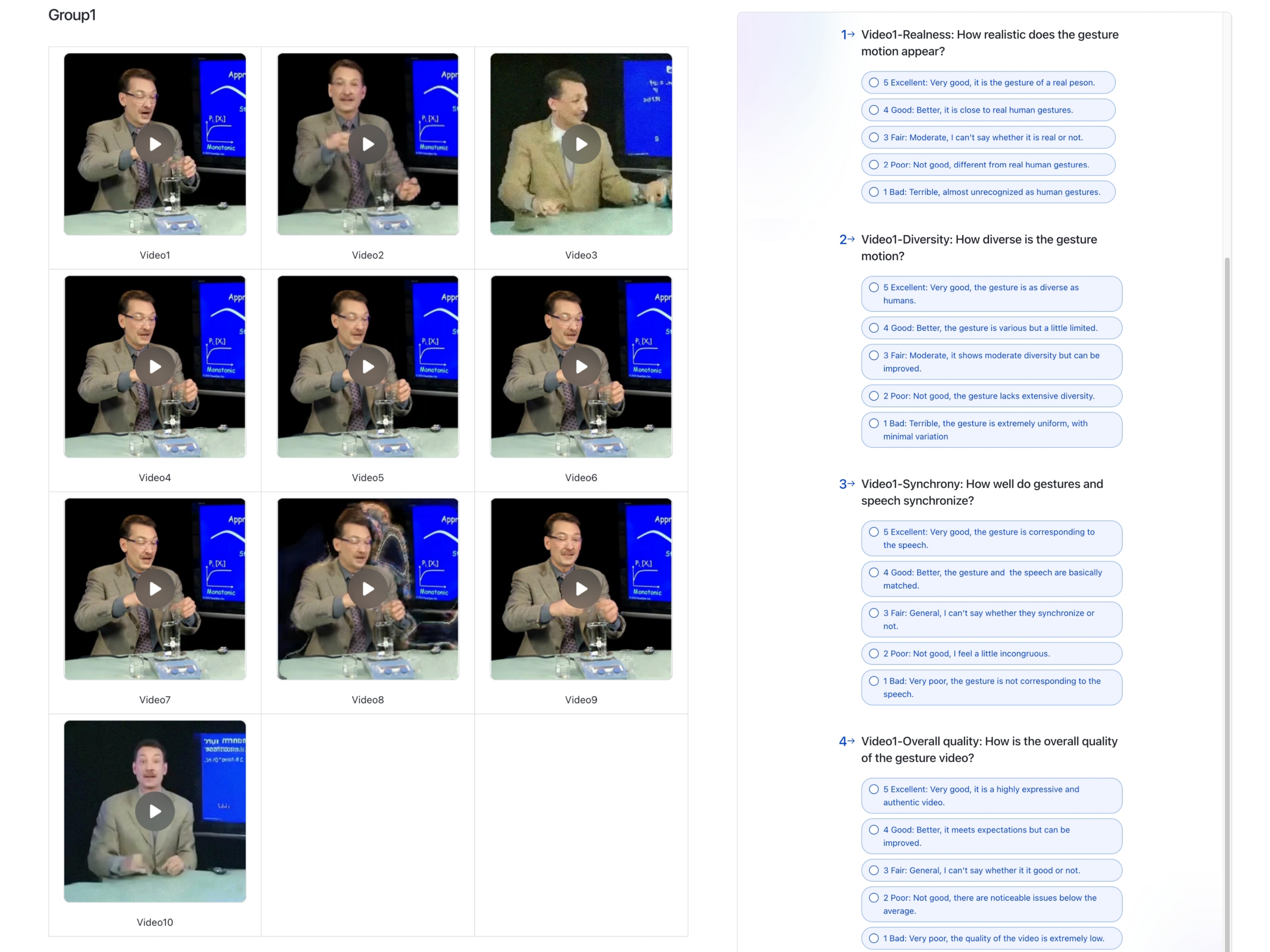}
    \caption{User study interface for comparison and ablation.}
    \label{fig:screen1}
  \end{subfigure}
  \hfill
  \begin{subfigure}{0.5\linewidth}
    \includegraphics[width=\linewidth]{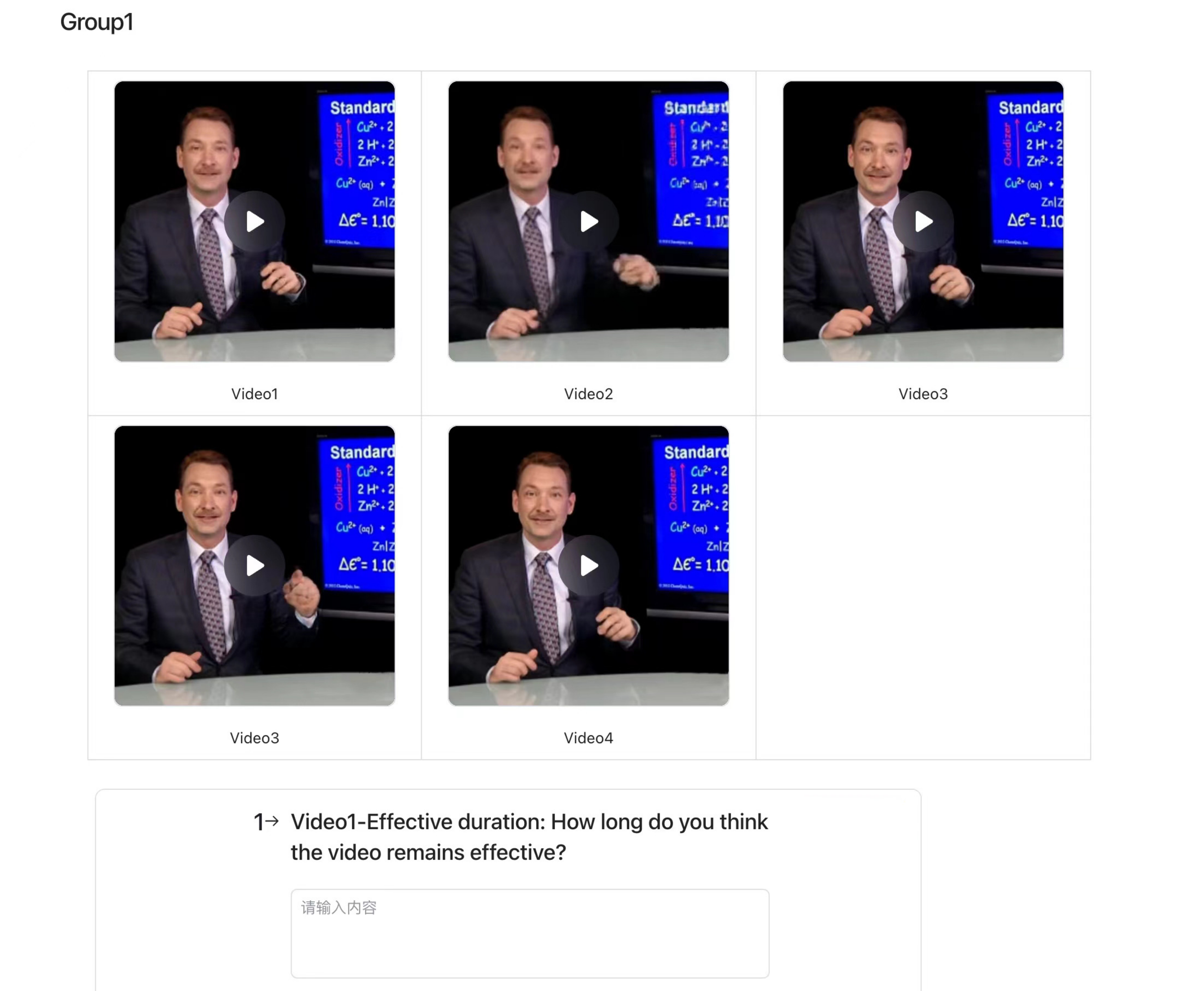}
    \caption{User study interface for rating effective duration.}
    \label{fig:screen2}
  \end{subfigure}
  \caption{Screenshots of the user study interface.}
  \label{fig:screen}
\end{figure*}

The user study is conducted by 20 participants with good English proficiency, involving 15 males and 5 females.
Each participant is remunerated about 15 USD for a rating of 40-50 minutes, which is approximately at the average wage level~\cite{yoon2022genea}. Screenshots of the rating interface used for comparison, the ablation study, and the evaluation of long video generation are presented in \cref{fig:screen}.

\section{Robustness and Effectiveness of Objective Metrics}
\label{sec:supp-effective}
\begin{figure}[t]
  \centering
   \includegraphics[width=0.6\linewidth]{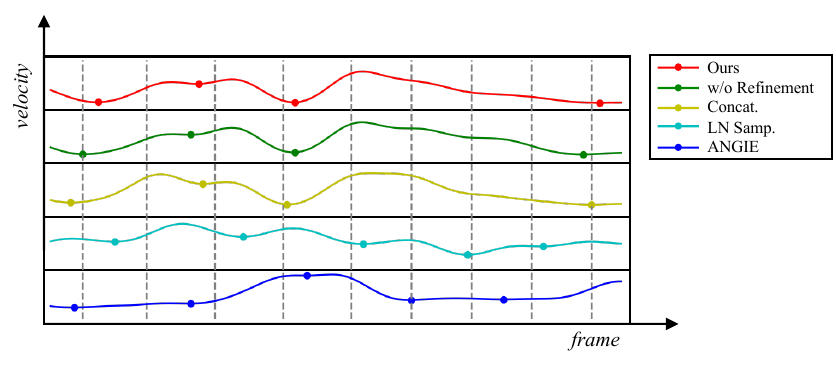}
   \caption{\textbf{Examples of velocity-frame curves of motion sequences} generated by each method for BAS analysis. For clear visualization, velocity is normalized and displayed without overlap. Dots represent gesture beats and dashed lines signify speech beats. ``Concat.'' is short for concatenation. ``LN Samp." is short for long noise sampling. ``LN Samp.'' and ``ANGIE" exhibit more gesture beats but are not aligned with speech beats.}
   \label{fig:bas_analysis}
\end{figure}

From the main paper, we observe: 1) ANGIE~\cite{angie} achieves higher BAS than ours. 2) Refinement brings lower BAS. 3) Sampling long noise and concatenation strategies have similar BAS. All these observations regarding BAS are inconsistent with subjective perceptions. Actually, BAS considers the distance between each audio beat with its nearest gesture beat, while gesture beats are defined as local velocity minima of 2D pose sequences filtered with a Gaussian kernel~\cite{bas}. In practice, we encounter unavoidable inter-frame jitters when extracting 2D poses for evaluation with the off-the-shelf pose estimator. Tremors such as those in ANGIE, blurred images without refinement, or almost stationary long noise sampling results could amplify the jitters of estimated poses and cause incorrect identification as denser gesture beats, reducing the distance between gesture and speech beats and thus incorrectly increasing BAS, which can be seen from \cref{fig:bas_analysis}. 
In summary, BAS is susceptible to unrelated factors, making it a less robust objective metric.
FGD, Diversity, and FVD are calculated in the feature space, making them somewhat more robust compared to BAS. 

Another interesting finding is that despite other metrics of our method being closer to the GT, FGD still exhibits a noticeable discrepancy. However, user study results strongly indicate the authenticity of our generated motion. One plausible explanation is that for FGD, we take the entire data, including the training and testing sets, as the real reference to calculate distribution distances. Given the rich diversity of gestures, there are inherent distribution gaps between the training and testing sets. Our model learns the data distribution from the training set, slightly deviating from the entire, while the GT of the testing set constitutes a portion of the overall distribution. This results in a noticeable difference in FGD. Referring to the training distribution reduces the difference (GT: 8.976 to 10.327 \vs ours: 18.131 to 13.285), providing supporting evidence.

Actually, previous studies~\cite{genea2020,kucherenko2023genea} indicate that co-speech gesture generation still lacks objective metrics perfectly consistent with human subjective perception. To summarize the above, we have to demonstrate that subjective evaluation remains the gold standard for co-speech gesture video generation just like any other technology in the field of human-machine interaction~\cite{genea2020}.

\section{Generalization Ability}
\label{sec:supp-generalization}
Gestures vary greatly between different speakers, so previous work typically trains an independent model for each person to capture individual styles. In contrast, we train a unified model jointly with the four speakers to ensure the scalability of our method. Experimental results indicate that even in this more challenging setting, our approach still generates gestures matching individual styles. Besides, we notice joint training brings about generalization ability to the speech of unseen speakers, which can be seen on our \href{https://github.com/thuhcsi/S2G-MDDiffusion}{homepage}. However, it is still hard to generalize to any given portrait at present. Yet, given two critical facts: 1) our method can animate unseen dressing appearances of the four given speakers, for the dataset contains various appearances of the same speaker, and 2) efforts like~\cite{animateanyone} on extensive multi-person datasets show stronger generalization ability to unseen portraits, we believe that our approach exhibits generalization potential, and a high-quality multi-speaker gesture video dataset may help to enhance it, which will be explored in our future work. 

\section{Time and Resource Consumption}
\label{sec:supp-time}
\cref{tab:consumption} indicates that our training and inference time are comparable to ANGIE~\cite{angie} and significantly shorter than MM-Diffusion~\cite{mmdiffusion}. Therefore, to the best of our knowledge, we achieve an optimal trade-off between time consumption and generation quality with distinct superiority in the latter. Although motion decoupling takes longer time, it greatly reduces the overall time and resource commitment compared to MM-Diffusion and other video generation works, \eg \cite{animateanyone} taking 14 days on 4 NVIDIA A100 GPUs for training\footnote{Experimental results from our reproduced code instead of official resources.}, providing a relatively efficient solution. Notably, our proposed diffusion model in the latent motion space achieves competitive generation results with relatively less time consumption, highlighting its necessity in the audio-to-motion process. Undeniably, repetitive diffusion denoising introduces extra inference time, and we will further explore methods like LCM~\cite{lcm} and Flow Matching~\cite{flowmatching} for acceleration. 

\begin{table}[!ht]
\centering
\caption{\textbf{Time consumption comparison} of training (6 NVIDIA A10 GPUs) and inference (1 NVIDIA GeForce RTX 4090 GPUs). }
\label{tab:consumption}
\resizebox{\columnwidth}{!}{%
\begin{tabular}{cccc}
\hline
Name & Training & Training Breakdown & \begin{tabular}[c]{@{}c@{}} Inference\\ (Generate a video of $\sim$10 sec) \end{tabular} \\
\hline
ANGIE & $\sim$5d & Motion Representation $\sim$3d + Quantization $\sim$0.2d + Gesture GPT $\sim$1.8d  & $\sim$30 sec\\
MM-Diffusion \ & $\sim$14d  & Generation $\sim$9d + Super-Resolution $\sim$5d& $\sim$600 sec \\
Ours & $\sim$5d & Motion Decoupling $\sim$3d + Motion Diffusion $\sim$1.5d + Refinement $\sim$0.5d & $\sim$35 sec\\
\hline
\end{tabular}%
}
\end{table}

\section{Limitations and Future Work}
\label{sec:supp-limit}
As research towards a relatively unexplored problem, there is still room for improvements in the following areas.

Despite significant superiority to existing methods, our generated videos still exhibit some accuracy issues of blurs and flickering, especially in hand details. This arises from the intricate structures of hands, characterized by varying movements like intersections and overlaps, which actually presents an unresolved challenge in the field of image and video generation~\cite{stablediffusion, animateanyone}. TPS-based motion decoupling effectively captures curved hand contours, making our method more adaptable to complex hand shapes than ANGIE~\cite{angie}, but still struggles to model structural details. The limited presence of hands in the frame drawing insufficient attention, coupled with the relatively weak inpainting capability of the image synthesis network, also leads to inaccurate hands. In addition, we observe that PATS dataset sourced from in-the-wild videos is of limited quality with noticeable hand motion blur, influencing the network's performance to some extent. Therefore, in our future work, we will: 1) refine our method, \eg prioritizing attention to hands and inpainting occlusion with more powerful pre-trained image generation models like SD model~\cite{stablediffusion}, and 2) collect high-quality gesture video data with clearer representations of hands to further enhance the generation quality.

Our current solution is unable to effectively synthesize the lip shape because there is a gap in the relationship between lips and gestures with speech. A unified framework for generating co-speech gestures and the lip shape simultaneously remains a valuable research problem, which we will explore in future work. In some showcases of the supplementary video, we use the off-the-shelf Wav2Lip~\cite{wav2lip} to synthesize lip shapes. Note that, the lip shape is not within the scope of this work, and generating lip shapes is just for better visual effects in the demo video.

For videos of bad quality, the accuracy of 2D poses from the pose estimator is compromised, leading to significant uncertainty when calculating all objective metrics regarding motion, especially BAS. Up until now, human subjective evaluation remains the most effective means of assessing generated gesture videos. Further exploration is needed to develop more robust and effective objective metrics.

\section{Dataset License}
\label{sec:supp-license}
We download the YouTube videos and perform preprocessing according to the video links in the metadata provided by the PATS dataset~\cite{pat1,pat2,pat3}. Video license ``CC BY - NC - ND4.0 International'' allows for non-commercial use. Although the video data includes personal identity information, we adhere to the data usage license, and our processed data, models, and results will be used only for academic purposes and not be permitted for commercial use.